\documentclass[10pt,twocolumn,letterpaper]{article}

\usepackage{cvpr}              

\usepackage{times}  
\usepackage{helvet}  
\usepackage{courier}  
\usepackage[hyphens]{url}  
\usepackage{graphicx} 
\usepackage{natbib}  
\usepackage{caption} 

\usepackage[table,xcdraw]{xcolor}

\usepackage{soul}
\usepackage{url}
\usepackage{cite}
\usepackage{amsmath,amssymb,amsfonts}
\usepackage{graphicx}
\usepackage{textcomp}

\usepackage{bm}

\usepackage{times}
\usepackage{soul}
\usepackage{url}

\usepackage{amsthm}
\usepackage{booktabs}
\usepackage[switch]{lineno}
\usepackage{mathtools}
\usepackage{times}  
\usepackage{helvet} 
\usepackage{courier}  
\usepackage{xparse}
\usepackage{booktabs}
\usepackage{xcolor}
\usepackage{url}            
\usepackage{amsfonts}       
\usepackage{nicefrac}       

\usepackage{rotating,multirow}

\usepackage{rotating,multirow}
\usepackage{amssymb} 
\usepackage{color}
\usepackage{lipsum}
\usepackage{tabularx}

\usepackage{helvet}  
\usepackage{courier}  

\usepackage{lscape}
\usepackage{makecell}
\usepackage{adjustbox}

\definecolor{cvprblue}{rgb}{0.21,0.49,0.74}
\usepackage[pagebackref,breaklinks,colorlinks,allcolors=cvprblue]{hyperref}

\def\paperID{******}
\def\confName{CVPR}
\def\confYear{2025}

\title{AlphaFace: High Fidelity and Real-time Face Swapper Robust to Facial Pose}

\author{Jongmin Yu$^{1,2}$, Hyeontaek Oh$^{1}$, Zhongtian Sun$^{3}$ Angelica I Aviles-Rivero$^{4}$\\ Moongu Jeo$^{5}$, and Jinhong Yang$^{1,6}$\\
ProjectG.AI$^{1}$, University of Cambridge$^{2}$, University of Kent$^{3}$, Tsinghua University$^{4}$\\Gwangju Institute of Science and Technology$^{5}$, Inje University$^{6}$\\
{\tt\small jy522@projectg.ai$^{1,2}$}
}

\begin{document}
\maketitle
\begin{abstract}
Existing face-swapping methods often deliver competitive results in constrained settings but exhibit substantial quality degradation when handling extreme facial poses. To improve facial pose robustness, explicit geometric features are applied, but this approach remains problematic since it introduces additional dependencies and increases computational cost. Diffusion-based methods have achieved remarkable results; however, they are impractical for real-time processing. We introduce AlphaFace, which leverages an open-source vision-language model and CLIP image and text embeddings to apply novel visual and textual semantic contrastive losses. AlphaFace enables stronger identity representation and more precise attribute preservation, all while maintaining real-time performance. Comprehensive experiments across FF++, MPIE, and LPFF demonstrate that AlphaFace surpasses state-of-the-art methods in pose-challenging cases. The project is publicly available on \url{https://github.com/andrewyu90/Alphaface_Official.git}.

\end{abstract}

\begin{figure*}[h!]
\centering
\includegraphics[width=0.9\linewidth]{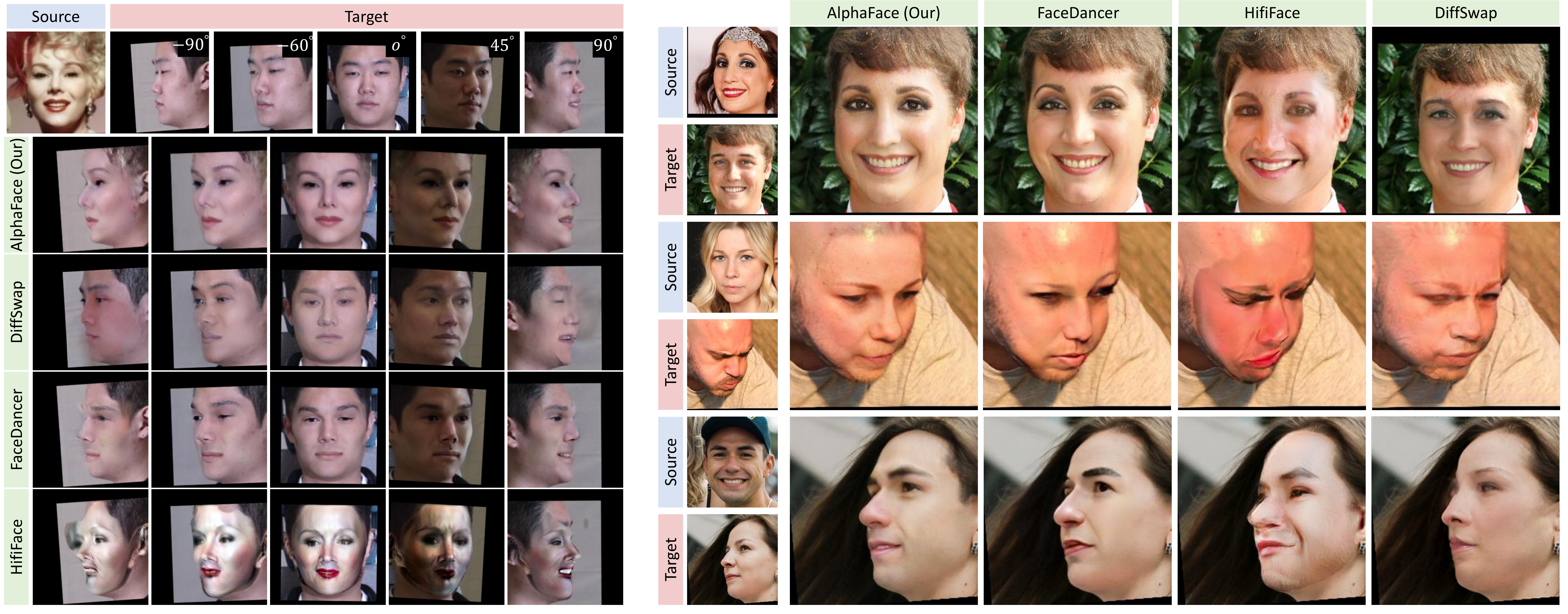}
\caption{\small Examples of the results of face identity swapping on various facial poses obtained by AlphaFace and recent SOTA methods based on diffusion model \citep{zhao2023diffswap} and exploit explicit geometric features~\citep{rosberg2023facedancer,wang2021hififace}. Compared to the frontal face image, the swapped results of the SOTA methods for extreme poses (greater than $\pm$45 degrees) remain highly distorted.}
\label{fig1}
\vspace{-2ex}
\end{figure*}

\section{Introduction}
Face swapping is the process of replacing an individual's facial identity in an image or video with another person's. Face swapping has applications in the entertainment and creative industries \citep{liu2023deepfacelab,waseem2023deepfakereview}, but it also raises ethical concerns, such as identity misuse and non-consensual content \citep{director2018submission}. Nevertheless, advancing face swapping remains technically crucial, including for Deepfake detection \citep{malik2022deepfakedetectionsurvey,mubarak2023deepfakeimpactsurvey}.

The central challenge in identity swapping is accurately transferring the source identity while preserving non-identity-related attributes of the target image (\eg, lighting, hairstyle, facial accessories, and facial poses) \citep{shiohara2023blendface}. Deep learning-based approaches have made significant progress \citep{chen2020simswap, chen2023simswap++, wang2024efficient}. However, most current systems are only effective for some controlled facial poses, which are inadequate for media content with complex motion dynamics or real-time situations that require high robustness to facial poses. Diffusion-based methods have demonstrated unprecedented photorealism  \citep{zhao2023diffswap,baliah2025realistic}. Yet, their high computational cost makes them unsuitable for interactive or real-time applications. Additionally, as shown in Figure \ref{fig1}, existing diffusion-based methods are still intractable for generating high-quality swapped faces under extreme facial poses. 
 
Those disruptions caused by significant angular variations in facial poses severely distort facial geometry, introduce self-occlusions, and disrupt the boundary alignment necessary for generating clean swapped faces. Although several strategies attempt to mitigate this issue using geometric priors or 3D supervision \citep{li20233d,wang2021hififace,rosberg2023facedancer}, as shown in Figure~\ref{fig1}, those methods still fail under extreme poses and introduce distortions, degrading image fidelity. Consequently, robust face identity swapping across facial poses remains a very challenging issue. 

In this work, we propose AlphaFace, a real-time face-swapping method that is robust to significant facial pose variations. AlphaFace achieves real-time performance by adopting a competitive architectural design that combines conventional Generative Adversarial Network (GAN)-like or autoencoder-like pipelines \citep{nirkin2019fsgan,chen2020simswap,li2019faceshifter}. Unlike other methods \citep{rosberg2023facedancer,wang2021hififace,li20233d} that exploit explicit geometric features of faces, AlphaFace enhances its semantic understanding by tightly integrating with strong semantic supervision from a large-scale vision-language model (VLM). We generate virtual text descriptions of facial images using a VLM and use this information to train AlphaFace with CLIP image and text encoders \citep{radford2021learning} and contrastive learning. 

Extensive experiments on FF++, MPIE, and LPFF demonstrate that AlphaFace surpasses recent state-of-the-art (SOTA) methods, particularly in extreme pose scenarios. On FF++, AlphaFace achieves a 98.77 identity (ID) retrieval score, a 1.24 pose error, and a 2.03 expression error. Similarly, on MPIE, it achieves the best cosine similarity score (0.471) and the lowest pose error (2.97) and expression error (3.03). FaceDancer \citep{rosberg2023facedancer} achieves 98.84, the highest ID retrieval score on FF++, but it performs worse on pose and expression errors. It also takes 78.3 ms per image, which is much slower than AlphaFace's 24.1 ms per image. These results highlight the benefits of leveraging VLM for feeding rich semantic supervision for improving the robustness of facial poses.

Our contributions are summarised as follows:

\begin{itemize}
  \item We introduce AlphaFace, a novel face-swapping framework that leverages rich semantic supervision from a large-scale VLM. By leveraging VLM-generated attribute text and CLIP encoders, AlphaFace applies image- and text-contrastive objectives, enabling robustness under extreme poses without explicit facial geometry priors.
\item We design an improved identity injection module, called `cross-adaptive identity injection' (CAII), that focuses on identity representation, isolating it from unnecessary attributes. 
\item We establish AlphaFace as a strong, open-source, and practical baseline. Through extensive experiments, we demonstrate state-of-the-art performance across identity, pose, expression, and fidelity metrics. We provide new insights into the role of a VLM-generated supervision in face-swapping.
\end{itemize}


\begin{figure*}[!t]
\centering
\includegraphics[width=0.95\linewidth]{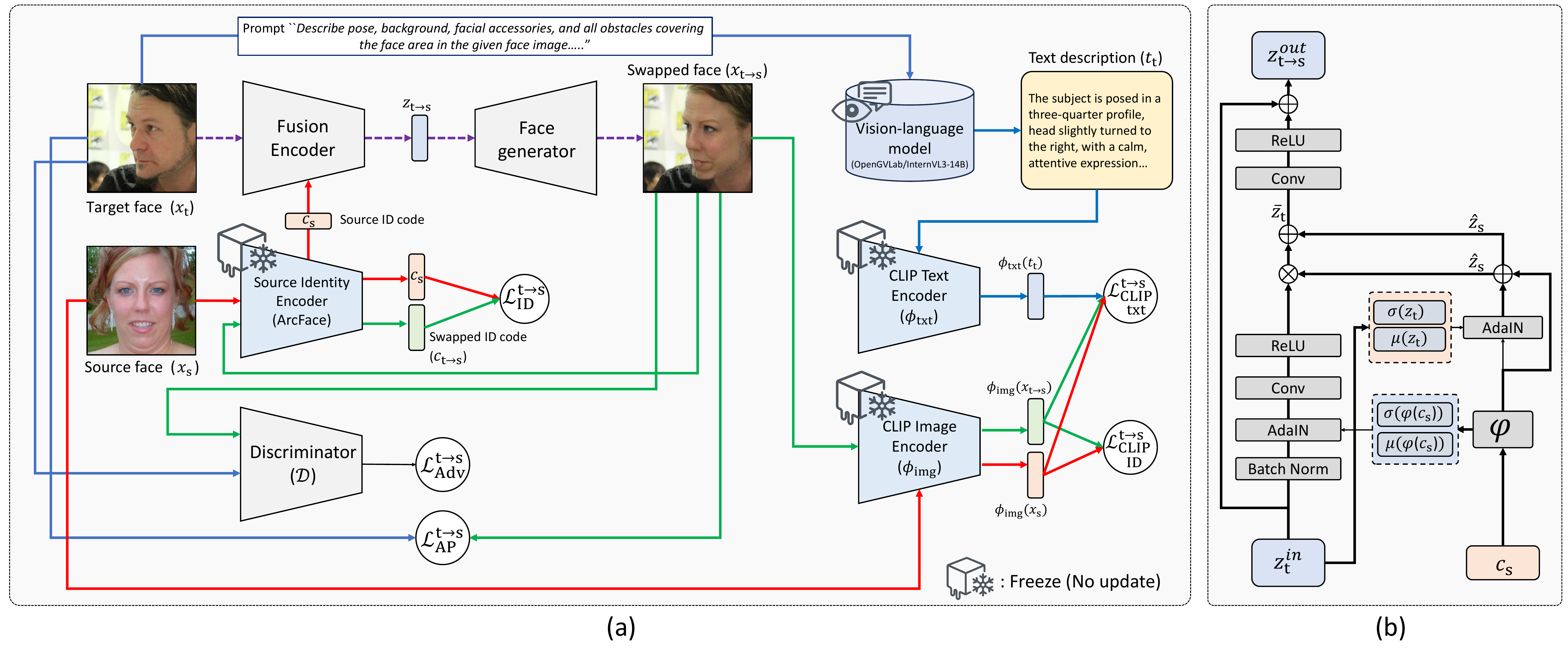}
\vspace{-1ex}
\caption{The detailed information for the architecture and workflow of AlphaFace. (a) illustrates the workflow details for training and testing of AlphaFace. (b) shows the architectural details of the cross-adaptive identity injection (CAII) block. The red, blue, and green arrow lines define the pipelines for source $x_{\text{s}}$, target $x_{\text{t}}$, and swapped face $x_{\text{t}\rightarrow{}\text{s}}$ images for training, respectively. The purple dotted lines define the pipeline to generate $x_{\text{t}\rightarrow{}\text{s}}$.}
\label{fig:workflow}
\vspace{-2ex}
\end{figure*}

\section{Related Work}
Face identity swapping has progressed rapidly with the advent of deep learning. The majority of methods rely on GANs and autoencoder-based frameworks, such as FaceShifter~\citep{li2019faceshifter}, FSGAN~\citep{nirkin2019fsgan}, and the SimSwap/SimSwap++~\citep{chen2020simswap,chen2023simswap++}, which achieved significant improvements in image realism compared with previous methods using heuristic algorithms \citep{bitouk2008face,dale2011video}. These approaches typically follow a two-step pipeline: (1) extract latent identity features from a source image using an independent encoder, and (2) merge these features with the target attribute representation to generate a swapped face.

Research within this paradigm has centred mainly on two objectives: improving source identity representation and preserving target-specific attributes (e.g., illumination, texture, accessories, and facial poses). Various advanced architectures, such as semantic-guided fusion layers~\citep{li2024learning} and identity injection blocks~\citep{chen2020simswap,chen2023simswap++}, and learning strategies~\citep{li2024learning,shiohara2023blendface} have been proposed to enhance source identity representation. Also, for preserving target attributes, many methods have presented various loss functions such as pixel-space reconstructions combined with perceptual losses~\citep{li2019faceshifter,liu2023deepfacelab,li2024learning,shiohara2023blendface}. The SimSwap series~\citep{chen2020simswap,chen2023simswap++} further introduced weakly supervised feature-matching losses, using adversarial discriminators as perceptual feature extractors, functioning similarly to VGG-based perceptual supervision.

Although the above methods achieved remarkable performance, the robustness of facial pose remains a critical challenge. Strategies for mitigating this issue often involve explicit geometric priors \citep{wang2021hififace,rosberg2023facedancer}. HiFiFace~\citep{wang2021hififace} incorporated 3D Morphable Models (3DMMs) to warp local textures. FaceDancer~\citep{rosberg2023facedancer} proposed interpretability-based regularisation for pose consistency. Diffusion-based methods~\citep{zhao2023diffswap,kim2025diffface,baliah2025realistic} have begun to challenge the field of face swapping. However, as shown in Figure \ref{fig1}, these approaches also do not guarantee good results and still struggle when the target face is significantly rotated or tilted, such as the cases of facial pose angle beyond $\pm{}$60$^{\circ}$.

In this work, we introduce AlphaFace, a real-time face-swapping framework that achieves higher fidelity and robustness to extreme facial poses. To ensure real-time performance, AlphaFace maintains the architectural compatibility of the GAN/autoencoder-based approaches. Instead of relying on explicit geometric features or advanced diffusion structures, AlphaFace utilises rich semantic information obtained from a vision-language model (VLM) during training. We generate text descriptions of face images using a VLM and apply them to the text and image encoders of the VLM to perform contrastive learning. The text and image encoders of CLIP \citep{radford2021learning} are used for contrastive learning. By leveraging rich semantic information extracted from a network trained on hundreds of thousands of text and image datasets, AlphaFace can explore face images under various conditions, thereby enhancing its fidelity and pose robustness without explicit facial geometry information.

\section{AlphaFace}
\label{sec:3}
\subsection{Architectural Details}
The AlphaFace framework is composed of three principal modules: 1) a source identity encoder, 2) a fusion encoder, and 3) a swapped face generator. Figure \ref{fig:workflow}(a) shows architectural details of the AlphaFace. Auxiliary components such as a discriminator and CLIP's text and image encoders are utilised during training but are not required at inference time; our description therefore focuses exclusively on the three essential modules.

The source identity encoder extracts discriminative latent features $c_{\text{s}}$ from the source face image $x_{\text{s}}$, ensuring that the generated output preserves the desired identity. To achieve robust and generalisable identity representation, we adopt ArcFace \citep{deng2019arcface} as the source identity encoder, following prior face-swapping work \citep{baliah2025realistic,chen2020simswap,chen2023simswap++,kim2025diffface,lin2012face,li2019faceshifter}.

The fusion encoder integrates $z_{\text{s}}$ from the source face with latent features $z_{\text{t}}$ from the target face $x_{\text{t}}$ for preserving attributes of the target face (\eg, pose, expression, illumination) while it shows source face identity. Its primary responsibility is the injection of source identity information into the combined latent representation. To integrate the target and source identity features, we present the \textit{Cross-Adaptive Identity Injection} (CAII) block. The CAII block is composed of two adaptive instance normalisation (AdaIN) \citep{huang2017arbitrary} and convolutional layers with residual operation. 

The CAII block firstly conducts batch normalisation (BN) when an input is given. After that, the CAII block applies AdaIN to normalise and re-style the probabilistic properties of the target latent feature $z_{\text{t}}$ using source identity code $c_{\text{s}}$, represented as follows:
\begin{equation}
 \label{eq:original_target}
\text{AdaIN}(z_{\text{t}},\varphi(c_{\text{s}})) = \sigma{}(\varphi(c_{\text{s}}))\frac{z_{\text{t}}-\mu(z_{\text{t}})}{\sigma(z_{\text{t}})}+\mu{}(\varphi(c_{\text{s}})),
\end{equation}
where $\mu{}$ and $\sigma{}$ denote functions to compute mean and standard deviation, $\varphi$ defines a neural network to map the source identity feature $c_{\text{s}}$ into a latent feature space taking the same dimensionality of $z_{\text{t}}$. $\varphi$ transform $c_{\text{s}}$ into a latent feature space which is more suitable for each block. The output of the AdaIN will go through an additional convolutional layer with rectified linear activation to obtain further abstracted features, represented by $\hat{z}_{\text{t}} = \alpha(\text{Conv}(\text{AdaIN}(z_{\text{t}},\varphi(c_{\text{s}}))))$, where $\alpha$ defines the rectified linear units. 

In addition to the AdaIN in target features, the CAII block applies AdaIN to source identity features, thereby attenuating the influence of information that is irrelevant to source identity representation and align it which is more suitable to the target latent features. After that, the normalised outputs are combined with the original values by applying residual operations, represented as follows:
\begin{equation}
 \label{eq:original_source}
\hat{z}_{\text{s}} = \text{AdaIN}(\varphi{}(c_{\text{s}}),z_{\text{t}})+\varphi{}(c_{\text{s}}),
\end{equation}
where $\hat{z}_{\text{s}}$ represents the final form of source identity features, which are actually used for identity injection. 

Identity injection is conducted by conducting element-wise multiplication and summation between $\hat{z}_{\text{s}}$ and $\hat{z}_{\text{t}}$, which is define by: $\bar{z}_{\text{t}} = (\hat{z}_{\text{t}}\otimes\hat{z}_{\text{s}})\oplus\hat{z}_{\text{s}}$, where $\otimes$ and $\oplus$ define the element-wise multiplication and summation, respectively. Figure \ref{fig:workflow}(b) illustrates the details of the entire architecture for the CAII block. This sequential design leverages the strengths of both normalisation strategies. Combined with the ID swapping loss, Eq. \ref{eq:original_target} encourages the swapped face to look more similar to the source identity; however, it also degrades the target attribute representation, which is observed by increasing reconstruction loss between target and swapped images during training \citep{li2024learning}. In this work, we apply Eq. \ref{eq:original_source}, which yields a target-adaptive source identity representation. 


Finally, the face generator reconstructs the swapped face by progressively upsampling the fused latent representation through a hierarchy of deconvolutional layers. This process produces high-resolution images that faithfully preserve the source identity while maintaining the target’s pose, expression, and surrounding context.

\subsection{Objective Functions and Learning}
\noindent
AlphaFace is trained by the objective functions combined with the five loss terms: 1) ID swapping loss, 2) attribute preserving loss, 3) adversarial learning loss, 4) and 5) CLIP-informed contrastive learning losses. The detailed explanation of those losses is as follows.

\noindent
\textbf{Loss for identity swap $\mathcal{L}^{\text{t}\rightarrow{}\text{s}}_{\text{ID}}$:} This encourages the swapped image $x_{\text{t}\rightarrow{}\text{s}}$ to have the same identity as $x_{\text{s}}$. Based on ArcFace $f_{\text{ID}}$, we extracted latent features from the swapped face $c_{\text{t}\rightarrow{}\text{s}}=f_{\text{ID}}(x_{\text{t}\rightarrow{}\text{s
}})$ and the source face image $c_{\text{s}}=f_{\text{ID}}(x_{\text{s}})$, respectively. The ID swapping loss is formulated based on cosine angular similarity using $c_{\text{t}\rightarrow{}\text{s}}$ and $c_{\text{s}}$, as follows:
\begin{equation}
 \label{eq:id}
\mathcal{L}^{\text{t}\rightarrow{}\text{s}}_{\text{ID}} = 1 - \frac{{c_{\text{s}}\cdot{}c_{\text{t}\rightarrow{}\text{s
}}}}{{{{\left\| c_{\text{s}} \right\|}_2}{{\left\| c_{\text{t}\rightarrow{}\text{s}} \right\|}_2}}}.
\end{equation}
where $\cdot$ denotes the dot product between $c_{\text{s}}$ and $c_{\text{t}\rightarrow{}\text{s}}$.

\noindent
\textbf{Loss for attribute preservation $\mathcal{L}^{\text{t}\rightarrow{}\text{s}}_{\text{AP}} $:} The above identity swap loss aims to improve the congruence of its identity with that of the source; the attribute preserving loss is focused on the faithful retention of identity-agnostic attributes such as illumination, cutaneous micro texture, and surrounding context. We define a target attribute-preserving loss that combines the two well-known losses defined in a pixel space and one loss defined in the latent space. 

The two loss functions defined in the pixel space are the masked reconstruction loss $\mathcal{L}^{\text{t}\rightarrow{}\text{s}}_{\text{Rec}}$ and the cyclic reconstruction loss $\mathcal{L}^{\text{t}\rightarrow{}\text{s}\rightarrow{}\text{t}}_{\text{Cycle}}$, formulated as follows:
\begin{equation}
\begin{aligned}
 \label{eq:recon}
    \mathcal{L}^{\text{t}\rightarrow{}\text{s}}_{\text{Rec}}(x_{\text{t}\rightarrow{}\text{s}},x_{\text{t}})   = \left\|\left( 1 - m_{\text{t}} \right) \otimes \left(  x_{\text{t}\rightarrow{}\text{s}}  -  x_{\text{t}} \right)  \right\|_1,
\end{aligned}    
\end{equation}
\begin{equation}
\begin{aligned}
 \label{eq:recon_cycle}
    \mathcal{L}^{\text{t}\rightarrow{}\text{s}\rightarrow{}\text{t}}_{\text{Cycle}}(x_{\text{t}\rightarrow{}\text{s}\rightarrow{}\text{t}},x_{\text{t}}) = \left\|    x_{\text{t}\rightarrow{}\text{s}\rightarrow{}\text{t}}  -  x_{\text{t}} \right\|_1, 
\end{aligned}    
\end{equation}
where $x_{\text{t}\rightarrow{}\text{s}}$ denotes the swapped image using the target image $x_{\text{t}}$ as the source image $x_{\text{s}}$, and $x_{\text{t}\rightarrow{}\text{s}\rightarrow{}\text{t}}$ indicates re-swapped results using the swapped image $x_{\text{t}\rightarrow{}\text{s}}$ as the target face and the target image $x_{\text{t}}$ as the source face. $m_{\text{t}}$ defines binary valued facial mask paired with $x_{\text{t}}$ obtained by Yu \etal \citep{yu2018bisenet}. 

$\mathcal{L}^{\text{t}\rightarrow{}\text{s}}_{\text{Rec}}$ is applied to explicitly learn visual information of the non-facial region of the target face image. Using $m_{\text{t}}$. However, strict spatial restrictions imposed by the binary mask may cause performance degradation because it does not provide all the necessary information to reconstruct the target appearance. In particular, to improve pose robustness, it is essential to provide precise information for the natural boundary of the face and the background, which a strict binary-valued mask cannot establish. $\mathcal{L}^{\text{t}\rightarrow{}\text{s}}_{\text{Cycle}}$ is used as a complementary term to address this issue. 

Additionally, we add a perceptual loss computed on deep features extracted from the VGG16 network: $\mathcal{L}_{\text{Percept}}^{\text{t}\rightarrow{}\text{s}}$. It supplies semantics-aware gradients that are robust to small shifts, directly align identity vectors, preserve fine textures, and regularise the generator against mode collapse. 

The total attribute-preserving loss function is defined by combining the above three loss terms, as follows:
\begin{equation}
\begin{aligned}
 \label{eq:cyclic}
    \mathcal{L}^{\text{t}\rightarrow{}\text{s}}_{\text{AP}}  = \mathcal{L}^{\text{t}\rightarrow{}\text{s}}_{\text{Rec}}+\mathcal{L}^{\text{t}\rightarrow{}\text{s}\rightarrow{}\text{t}}_{\text{Cycle}}+\mathcal{L}_{\text{Percept}}^{\text{t}\rightarrow{}\text{s}}.
\end{aligned}    
\end{equation}
The attribute-preserving loss maintains overall photometric consistency, while the perceptual loss guides the model to produce identity-faithful, sharp, and visually realistic swaps even under varying pose, illumination, or expression.

\noindent
\textbf{Adversarial learning loss $\mathcal{L}^{\text{t}\rightarrow{}\text{s}}_{\text{Adv}}$:} Adversarial objectives are routinely employed to elevate the visual fidelity of identity-swapped faces, chiefly by restoring high-frequency cues, such as edge acuity and fine textural details, which govern visual sharpness \citep{li2024learning, li2019faceshifter, chen2020simswap}. We utilise the PatchGAN \citep{henry2021pix2pix} to enhance the visual quality of the swapped face. The applied adversarial learning into multiple small-sized patches extracted from the single images, allowing it to regenerate more detailed high-frequency visual content. 

\noindent
\textbf{CLIP-informed contrastive learning:} Conventional objectives for face identity swapping (\eg, identity, reconstruction, and adversarial losses) may still yield distortions under extreme head poses (see Fig.~\ref{fig1}), even with architectural advances. To increase robustness, we augment training with supervisory signals from a foundation-scale VLM. Such models, trained on web-scale image-text corpora, provide semantically rich representations that complement single-domain encoders (\eg, ArcFace~\citep{deng2019arcface}), which are typically optimised on canonical, near-frontal portraits.

We address the above challenge by formulating two contrastive learning tasks using an open-source VLM and the image and text encoders of the CLIP. The first aligns the swapped image with a textual description of target attributes; the second enforces visual identity consistency between the swapped output and the source face.

First, CLIP-based image-to-text contrastive learning loss $\mathcal{L}^{\text{t}\rightarrow{}\text{s}}_{\text{CLIP-text}}$ is formulated as follows:
\begin{equation}
\begin{aligned}
\mathcal{L}^{\text{t}\rightarrow{}\text{s}}_{\text{CLIP-text}}=\tau{}\left(1-
\frac{
  \left\langle
    \phi_{\text{img}}\!\bigl(x_{\text{t}\rightarrow{}\text{s}}\bigr),\;
    \phi_{\text{text}}\!\bigl(t_{\text{t}}\bigr)
  \right\rangle
}{
  \lVert \phi_{\text{img}}\!\bigl(x_{\text{t}\rightarrow{}\text{s}}\bigr) \rVert\,
  \lVert \phi_{\text{text}}\!\bigl(t_{\text{t}}\bigr) \rVert
}\right),
\label{eq:clip_contrastive1}
\end{aligned}
\end{equation}
where $\phi_{\text{img}}$ and $\phi_{\text{text}}$ are the image and text encoders of CLIP. $t_{\text{t}}$ is the description of the target face image. The description is automatically obtained by an open-source large vision-language model (VLM). 

$\tau{}$ is an indicator to check whether given samples are valid for computing the loss or not, and the value of the function is set to 1 if it satisfies the following condition: $\frac{
  \left\langle
    \phi_{\text{img}}\!\bigl(x_{\text{t}}\bigr),\;
    \phi_{\text{text}}\!\bigl(t_{\text{t}}\bigr)
  \right\rangle
}{
  \lVert \phi_{\text{img}}\!\bigl(x_{\text{t}}\bigr) \rVert\,
  \lVert \phi_{\text{text}}\!\bigl(t_{\text{t}}\bigr) \rVert
} > \frac{
  \left\langle
    \phi_{\text{img}}\!\bigl(x_{\text{t}\rightarrow{}\text{s}}\bigr),\;
    \phi_{\text{text}}\!\bigl(t_{\text{t}}\bigr)
  \right\rangle
}{
  \lVert \phi_{\text{img}}\!\bigl(x_{\text{t}\rightarrow{}\text{s}}\bigr) \rVert\,
  \lVert \phi_{\text{text}}\!\bigl(t_{\text{t}}\bigr) \rVert
}$; otherwise, it sets 0. $\tau{}$ activates the loss only when the swapped image is less consistent with the target description than the original target (\ie, text-image similarity between $x_{\text{t}\rightarrow{}\text{s}}$ and $t_{\text{t}}$ is smaller than the one between $x_{\text{t}}$ and $t_{\text{t}}$), signaling an attribute mismatch that warrants correction.

Additionally, to reinforce source identity representation, we add the CLIP-based ID swapping loss,  
$\mathcal{L}^{\text{t}\rightarrow{}\text{s}}_{\text{CLIP-ID}}$, which is represented by:
\begin{equation}
\begin{aligned}
\mathcal{L}^{\text{t}\rightarrow{}\text{s}}_{\text{CLIP-ID}}=1-
\frac{
  \left\langle
    \phi_{\text{img}}\!\bigl(x_{\text{t}\rightarrow{}\text{s}}\bigr),\;
    \phi_{\text{img}}\!\bigl(x_{\text{s}}\bigr)
  \right\rangle
}{
  \lVert \phi_{\text{img}}\!\bigl(x_{\text{t}\rightarrow{}\text{s}}\bigr) \rVert\,
  \lVert \phi_{\text{img}}\!\bigl(x_{\text{s}}\bigr) \rVert
}.
\label{eq:clip_contrastive2}
\end{aligned} 
\end{equation}
By using the above CLIP-informed losses, we can feed richer semantic information during AlphaFace training, which consequently improves the visual quality and face pose robustness of the face identity-swapping model. We demonstrate the effectiveness of the two CLIP-based contrastive learning losses in our ablation study. 

\noindent
\textbf{Total objective:} The overall optimisation criterion is constructed by linearly combining the previously defined losses, each modulated by a dedicated balancing weight. It is expressed as
\begin{equation}
\begin{aligned}
    \mathcal{L}^{\text{t}\rightarrow{}\text{s}}_{\text{Total}} & = \lambda_{\text{ID}}\mathcal{L}^{\text{t}\rightarrow{}\text{s}}_{\text{ID}}+\lambda_{\text{AP}}\mathcal{L}^{\text{t}\rightarrow{}\text{s}}_{\text{AP}}+\lambda_{\text{Adv}}\mathcal{L}^{\text{t}\rightarrow{}\text{s}}_{\text{Adv}}\\ & +\lambda_{\text{CLIP}}(\mathcal{L}^{\text{t}\rightarrow{}\text{s}}_{\text{CLIP-text}}+\mathcal{L}^{\text{t}\rightarrow{}\text{s}}_{\text{CLIP-ID}}),
\end{aligned}
\end{equation}
where $\lambda_{\text{ID}}$, $\lambda_{\text{AP}}$, $\lambda_{\text{Adv}}$, and $\lambda_{\text{CLIP}}$ define balancing weights.


\section{Experiments}
\label{sec:4}
\subsection{Dataset and Experimental Settings}
\noindent
\textbf{Dataset curation:} For training, we employ VGGFace2-HQ \citep{cao2018vggface2} and CelebA-HQ \citep{karras2018progressive}. Ablations and comparative evaluations are conducted on FaceForensics++ (FF++) \citep{rossler2019faceforensics++}, Multi-Pose, Illumination, Expressions (MPIE) \citep{gross2010multi}, and Large-Pose Flickr Face (LPFF) \citep{wu2023lpff}. Although FF++ is the de facto benchmark in face-swapping research, its design does not explicitly stress pose diversity. MPIE and LPFF, which target wide yaw/pitch variations, therefore complement FF++ and enable a more rigorous assessment of AlphaFace concerning pose robustness. 

We conduct data-preprocessing applied to most of the face swapping methods \citep{chen2020simswap,chen2023simswap++,shiohara2023blendface,wang2021hififace,rosberg2023facedancer,li2019faceshifter} to reduce the influence of visual quality for the experimental results. The resolutions of $x_{\text{s}}$ and $x_{\text{t}}$ are regularised to $112\times{}112$ and $256\times{}256$. We provide detailed information on the above datasets and the pre-processing in the Appendix \ref{appx:datasets}.

OpenGVLab/InternVL3-14B \citep{chen2024internvl}, an open-source VLM, is used to generate $t_{\text{t}}$. The prompt is \textit{``Describe pose, background, facial accessories, and all obstacles covering the face area in the given face image. Only 70 words are allowed."}. It is intractable to verify all text descriptions in a large-scale training dataset manually; therefore, we do not verify their validity. Appendix \ref{appx:pair} shows some examples of our training samples.

\noindent
\textbf{Evaluation metrics and protocol:} We mainly quantified performance using four well-established criteria for face identity swapping: identity proximity measured by cosine similarity (CSIM), identity-retrieval accuracy, and the pose and expression errors, which are two attribute metrics capturing pose alignment and expression matching. In addition, we utilise Frechet Inception Distance (FID) to evaluate the fidelity of face swapping methods. 

For FF++, we adhered to the evaluation procedures described by Li \etal \citep{li2019faceshifter} and Chen \etal \citep{chen2020simswap}, enabling fair comparisons. For MPIE and LPFF, no widely accepted quantitative benchmark exists; accordingly, our primary analysis on these datasets is qualitative, with extensive visualisations. Given that MPIE provides multiple images per identity, similar to FF++, it is possible to conduct a controlled protocol presented from FF++. We randomly picked 1,000 source faces from the CelebA-HQ and treated every MPIE sample as a target. After that, we conduct face swapping and compute the pose and expression errors. CSIM is used as an alternative to the identity-retrieval accuracy.



As relatively few works reported results on MPIE and LPFF, we benchmark against methods with publicly released implementations: FSGAN \citep{nirkin2019fsgan}, SimSwap \citep{chen2020simswap}, BlendFace \citep{shiohara2023blendface}, HifiFace \citep{wang2021hififace}, DiffSwap \citep{zhao2023diffswap}, and FaceDancer \citep{rosberg2023facedancer}.  Appendix~\ref{appx:repositories} provides access information. 

\noindent
\textbf{Implementation details:} The balancing weights are set by 10.0, 0.5, and 1.0 for $\lambda_{\text{ID}}$, $\lambda_{\text{AP}}$,  $\lambda_{\text{Adv}}$, respectively. We set 1.0 for $\lambda_{\text{CLIP}}$. The batch size is decided to be 8. Adam Optimiser is employed. The initial learning rate is set to 0.01, and it is decayed every five epochs by multiplying by 0.9. The total epoch is set to 50. We used two A6000 GPUs for training and one RTX 4090 for testing, respectively.

\begin{table}[t]
\centering\resizebox{\linewidth}{!}{
\begin{tabular}{l|c|c|c|c}
\toprule
\multicolumn{5}{c}{Experimental results on FF++ dataset} \\
\midrule
 \textbf{Objective setting} & \textbf{ID retrieval $\uparrow$ } & \textbf{pose error $\downarrow$} & \textbf{expr error $\downarrow$}  & \textbf{FID $\downarrow$} \\
\midrule
${\text{w/o-CLIPs}}$ & 96.82 &  2.75 & 3.82 & 4.95 \\
\midrule
${\text{CLIP-w/o-text}}$  & 97.67 &  2.07 & 2.58  & 2.90\\
\midrule
${\text{CLIP-w/o-ID}}$  & 98.52 &  1.58 & 2.19  & 3.12 \\
\midrule
${\text{w-CLIPs}}$  & \cellcolor[HTML]{D7FCD7}\textbf{98.77} &  \cellcolor[HTML]{D7FCD7}\textbf{1.24} & \cellcolor[HTML]{D7FCD7}\textbf{2.03}   & \cellcolor[HTML]{D7FCD7}\textbf{2.71} \\ 
\midrule
\multicolumn{5}{c}{Experimental results on MPIE dataset} \\
\midrule
\textbf{Objective setting} & \textbf{CSIM $\uparrow$ } & \textbf{pose error $\downarrow$} & \textbf{expr error $\downarrow$} & \textbf{FID $\downarrow$}\\
\midrule
${\text{w/o-CLIPs}}$   & 0.427 &  4.19 & 5.03  & 11.04\\
\midrule
${\text{CLIP-w/o-text}}$   & 0.465 &  3.82  &3.43  & 8.12\\
\midrule
${\text{CLIP-w/o-ID}}$   & 0.467 &  3.12 & 3.17  & 9.61 \\
\midrule
${\text{w-CLIPs}}$  & \cellcolor[HTML]{D7FCD7}\textbf{0.471} &  \cellcolor[HTML]{D7FCD7}\textbf{2.97} & \cellcolor[HTML]{D7FCD7}\textbf{3.03}  &  \cellcolor[HTML]{D7FCD7}\textbf{7.78} \\ 
\bottomrule
\end{tabular}
}
\caption{The quantitative results regarding CLIP-informed losses (Eq. \eqref{eq:clip_contrastive1} and Eq. \eqref{eq:clip_contrastive2}) of the AlphaFace. $\text{w}$ and $\text{w/o}$ stand for 'with' and `without'; so that ${\text{CLIP-w/o-text}}$ defines the AlphaFace trained without the CLIP-based image-to-text contrastive learning $\mathcal{L}^{\text{t}\rightarrow{}\text{s}}_{\text{CLIP-text}}$. \colorbox[HTML]{D7FCD7}{\textbf{Green}} highlights the best performances.}
\label{table:abl}
\vspace{-2ex}
\end{table}

\subsection{Ablation studies on CLIP-based losses}
Table~\ref{table:abl} presents the quantitative evaluation results comparing two architectural variants of AlphaFace on the FF++ and MPIE datasets. We evaluate four configurations: w/o-CLIP (without the CLIP-informed contrastive learning), CLIP-w/o-text ($\mathcal{L}^{\text{t}\rightarrow{}\text{s}}_{\text{CLIP-ID}}$ only), CLIP-w/o-ID ($\mathcal{L}^{\text{t}\rightarrow{}\text{s}}_{\text{CLIP-text}}$ only), and w-CLIPs ($\mathcal{L}^{\text{t}\rightarrow{}\text{s}}_{\text{CLIP-ID}}+\mathcal{L}^{\text{t}\rightarrow{}\text{s}}_{\text{CLIP-text}}$). Metrics comprise ID retrieval/CSIM (higher is better) and pose/expression errors and FID (lower is better). The w-CLIPs yields the strongest overall performance, simultaneously enhancing identity fidelity and reducing pose and expression errors.

On the experimental results using the FF++, relative to w/o-CLIP (96.82 ID retrieval score, 2.75 pose error, 3.82 expression error, and 4.95 FID), w-CLIPs achieves obviously better performances, which are 98.77 ID retrieval score, 1.24 pose error, 2.03 expression error, and 2.71 FID. Interestingly, the $\mathcal{L}^{\text{t}\rightarrow{}\text{s}}_{\text{CLIP-text}}$ is the more potent single loss: CLIP-w/o-ID delivers 98.52 ID retrieval score, 1.58 pose error, and 2.19 expression error, outperforming CLIP-w/o-text (97.67, 2.07, and 2.58, respectively) on all three metrics. The trend also repeats in the results using the MPIE. The w-CLIPs produces the best performances, which are 0.471 CSIM, 2.97 pose error, and 3.03 expression error. 

These results indicate that textual supervision supplies identity-agnostic semantic constraints that better suppress pose/expression errors whilst preserving source identity. In contrast, CLIP-based identity-swapping loss alignment remains valuable for additional geometric/appearance tethering, even though a basic ID-swap loss (Eq. \eqref{eq:id}) is already applied. Improvements from the two CLIP losses are complementary but sub-additive, consistent with partially overlapping supervisory signals.



\begin{table}[t]
\centering\resizebox{\linewidth}{!}{
\begin{tabular}{l|c|c|c|c}
\toprule
\multicolumn{5}{c}{Experimental results on FF++ dataset} \\
\midrule
 \textbf{ID-injection setting} & \textbf{ID retrieval $\uparrow$ } & \textbf{pose error $\downarrow$} & \textbf{expr error $\downarrow$}  & \textbf{FID $\downarrow$}  \\
\midrule
Unidirectional (w/o-Eq. \ref{eq:original_source})  & \cellcolor[HTML]{D7FCD7}\textbf{98.80} &  1.27 & 2.68 & 5.27 \\
\midrule
CAII (Our)  & 98.77 &  \cellcolor[HTML]{D7FCD7}\textbf{1.24} & \cellcolor[HTML]{D7FCD7}\textbf{2.03}  & \cellcolor[HTML]{D7FCD7}\textbf{2.71}  \\ 
\midrule
\multicolumn{5}{c}{Experimental results on MPIE dataset} \\
\midrule
 \textbf{ID-injection setting} & \textbf{CSIM $\uparrow$ } & \textbf{pose error $\downarrow$} & \textbf{expr error $\downarrow$} \\
\midrule
Unidirectional (w/o-Eq. \ref{eq:original_source})   & 0.452 &  3.41 & 4.18 & 10.9 \\
\midrule
CAII (Our)  & \cellcolor[HTML]{D7FCD7}\textbf{0.471} &  \cellcolor[HTML]{D7FCD7}\textbf{2.97} & \cellcolor[HTML]{D7FCD7}\textbf{3.03} & \cellcolor[HTML]{D7FCD7}\textbf{7.78}  \\ 
\bottomrule
\end{tabular}
}
\caption{The quantitative results with respect to the identity injection approaches. We compare the cross-adaptive identity injection (CAII) with unidirectional identity injection, which is commonly used for the face identity swapping literature. \colorbox[HTML]{D7FCD7}{\textbf{Green}} highlights the best performances.}
\label{table:abl2}
\vspace{-2ex}
\end{table}

\subsection{Ablation studies on identity injection}
One of the major engineering differences between the AlphaFace and existing face-swapping methods \citep{li2019faceshifter,chen2020simswap,chen2023simswap++,rosberg2023facedancer,wang2021hififace} is the CAII block. The CAII applies additional alignment to the source identity using target latents before injecting it into the target latents; thereby, it mitigates the adverse effect of irrelevant information in the source identity code on the representation of source face identity, and generates a more well-aligned source identity code for target latent features. In our ablation study, we compare the performance of AlphaFace compiled with CAII and a unidirectional identity injection block. The uni-directional identity injection block is defined by skipping the Eq. \ref{eq:original_source}. The defined uni-directional identity injection is analogous to the identity injection approaches of SimSwap \citep{chen2020simswap}, MegaFS \citep{zhu2021one}, and Blendface \citep{shiohara2023blendface}.

\begin{figure*}[t]
\centering
\includegraphics[width=0.80\textwidth]{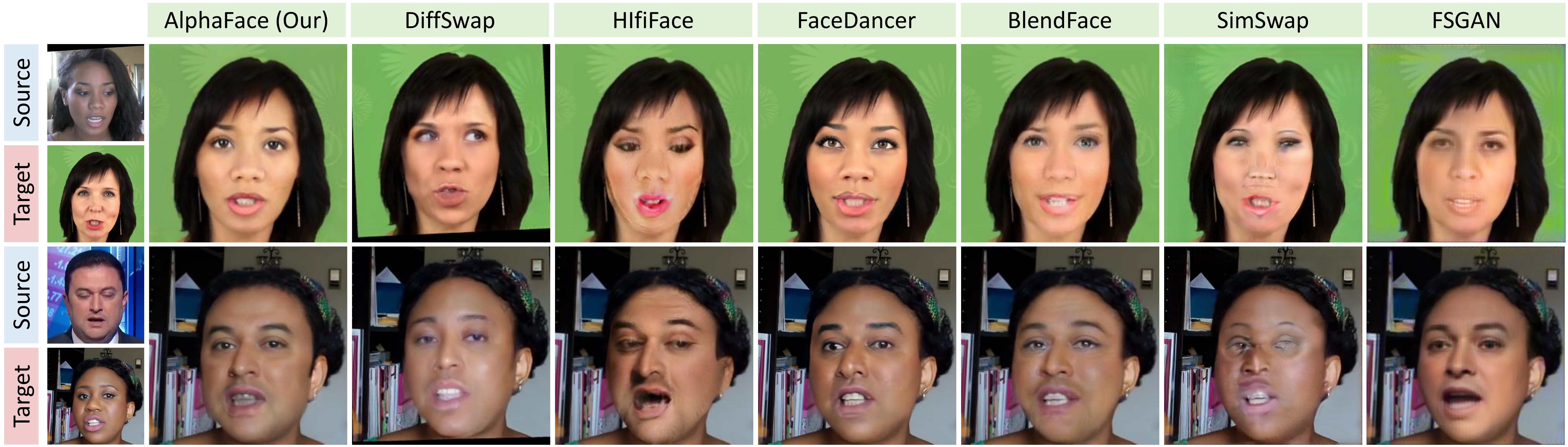}
\caption{Qualitative results of AlphaFace and the existing SOTA methods \citep{shiohara2023blendface,nirkin2019fsgan,chen2020simswap,wang2021hififace,rosberg2023facedancer,zhao2023diffswap} on FF++ dataset \citep{rossler2019faceforensics++}. The extended results are shown in Appendix \ref{appx:ff++_results}.}
\label{fig:ff_qualtitative}
\vspace{-2ex}
\end{figure*}

Table \ref{table:abl2} presents the quantitative results depending on the identity injection approaches, respectively. On FF++, identity retrieval is 98.80 in the unidirectional setting and 98.77 in CAII, showing that changing the injection strategy does not reduce identity preservation in practice. At the same time, CAII lowers pose error from 1.27 to 1.24, reduces expression error from 2.68 to 2.03, and improves FID from 5.27 to 2.71, indicating better alignment with target pose/expression and better visual quality under the same identity level. On MPIE, the advantages are more pronounced: CSIM increases from 0.452 to 0.471, pose error decreases from 3.41 to 2.97, expression error decreases from 4.18 to 3.03, and FID improves from 10.9 to 7.78. Since MPIE involves larger pose and illumination variations, these results suggest that CAII handles them more stably than the unidirectional scheme while still maintaining high identity similarity. The corresponding qualitative results are provided in Appendix \ref{appx:abl_connection}.


\subsection{Performance Comparison}

\noindent
\textbf{Comparison on FF++:} To preclude the possibility that the CLIP-based supervision simply privileges extreme head-pose cases, we evaluate AlphaFace on widely used benchmarks against strong state-of-the-art (SOTA) baselines. Table~\ref{Table:ff_results} reports results on FF++ under five criteria: 1) identity-retrieval accuracy, 2) pose error, 3) expression error, 4) FID, and 5) execution speed. Figure~\ref{fig:ff_qualtitative} provides representative qualitative exemplars.

Among all methods, AlphaFace offers the most harmonised performance-efficiency trade-offs across identity, pose, expression, FID, and processing speed. Although FaceDancer attains the highest ID retrieval scores (98.84), it incurs substantially larger expression errors with a slow runtime of about 78.3 ms per image (approximately 12.8 frames per second (FPS)), which is intractable for real-time applications. DiffSwap achieves an ID accuracy of 98.54, with pose/expression errors of 2.45 and 5.35, respectively. It produces the best FID performance, which is 2.16. However, it takes approximately 46 seconds per image, making it impractical for real-time performance. In contrast, AlphaFace achieves markedly lower pose/expression errors (1.24 and 2.03) with an inference time of 24.1 ms, while maintaining a competitive identity preservation (98.77 ID retrieval score) and a good fidelity (2.71). As a result, these results demonstrate that AlphaFace is not only robust in preserving target geometry and expressions but is also operationally suited to real-time face identity swapping without sacrificing quantitative performance.

\begin{figure*}
\centering
\begin{subfigure}{0.85\textwidth}
    \centering
    \includegraphics[width=\textwidth]{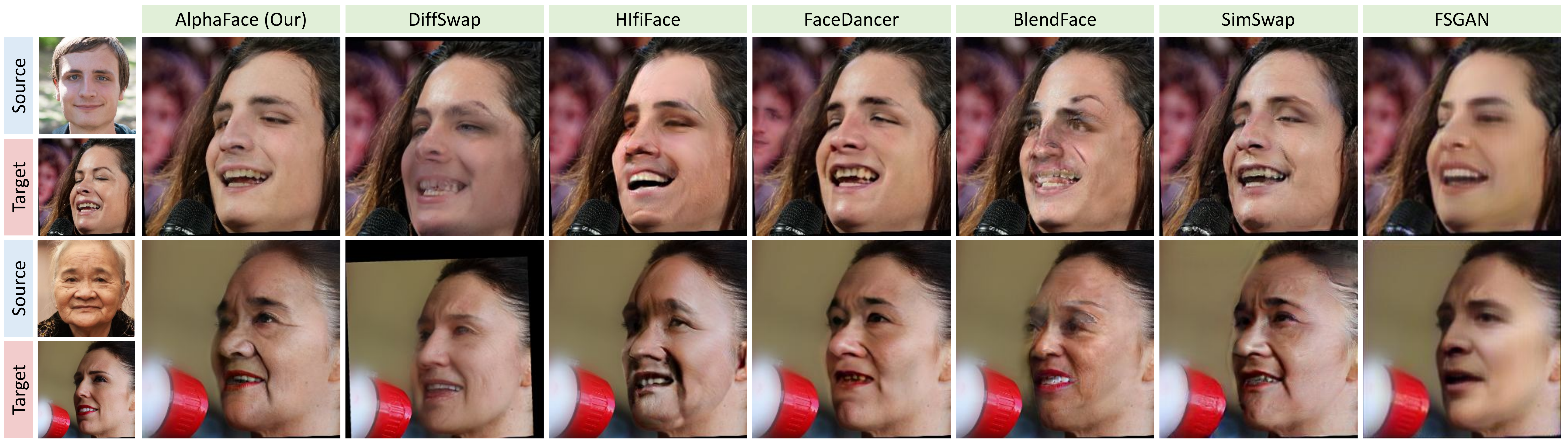}
    \caption{ }
\end{subfigure}
\hfill
\begin{subfigure}{0.85\textwidth}
    \centering
    \includegraphics[width=\textwidth]{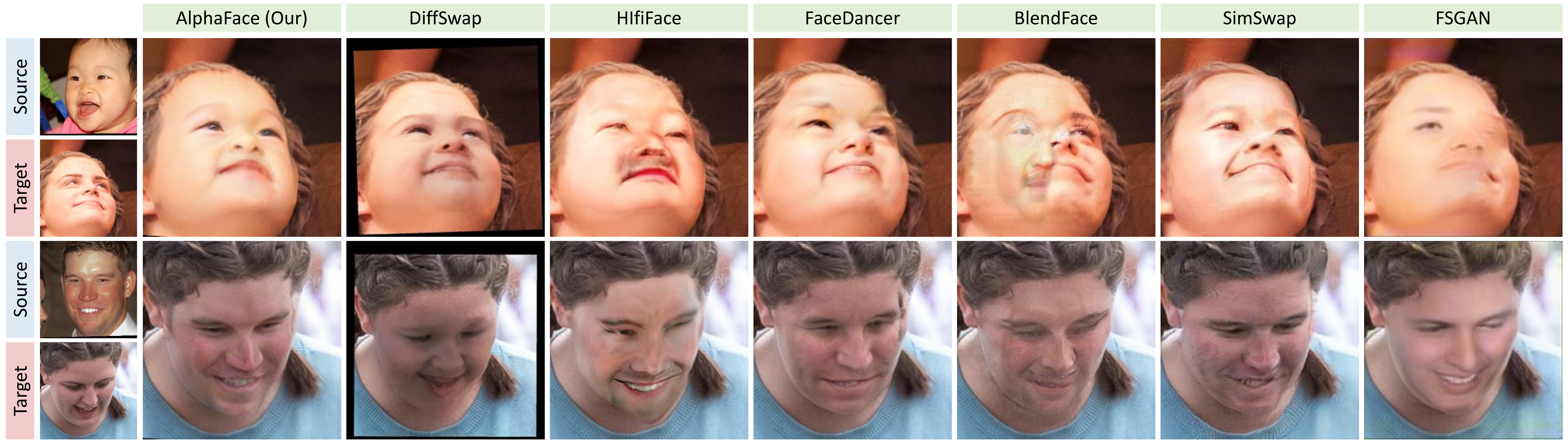}
    \caption{ }
\end{subfigure}
\caption{Face swapping results of the AlphaFace and other existing SOTA methods \citep{zhao2023diffswap,wang2021hififace,rosberg2023facedancer,shiohara2023blendface,chen2020simswap,nirkin2019fsgan} on the LPFF dataset \citep{wu2023lpff}. (a) and (b) represents the swapping results on rotated and tilted facial pose cases, respectively. Extended results are provided in Appendix \ref{appex:lpff_extended}.}
\label{fig:lpff_results}
\vspace{-2ex}
\end{figure*}

\begin{table}[t!]
\centering\resizebox{\columnwidth}{!}{
\begin{tabular}{l|c|c|c|c|c}
\toprule
\textbf{Method} & \textbf{ID ret $\uparrow$ } & \textbf{Pose err $\downarrow$} & \textbf{Expr err $\downarrow$} & \textbf{FID $\downarrow$} &  \textbf{Speed (ms) $\downarrow$}\\
\midrule
FaceSwap~\citep{rossler2019faceforensics++}& 72.69 & 2.58 & 2.89 & - & - \\
DeepFakes~\citep{DeepFakes} &  88.39 & 4.64 & 3.33 & -  & -\\
FaceShifter~\citep{li2020advancing}&  90.68 & 2.55 & 2.82 & - & - \\
MegaFS~\citep{zhu2021one}&   90.83 & 2.64 & 2.96 &-  & - \\
FSLSD~\citep{xu2022high}&  90.05 & 2.46 & 2.79& -  & -\\
RAFSwap~\citep{xu2022region}&   92.54 &3.21 & 3.60 & - & - \\
FaceSwapper~\citep{li2024learning} & 94.48 & 2.10 & 2.69 & -  & - \\
\hline
FSGAN$^{\dagger}$~\citep{nirkin2019fsgan}&   61.07 & 3.31 & 3.02 & 15.36  & \cellcolor[HTML]{D7FCD7}\textbf{21.5} \\
SimSwap$^{\dagger}$~\citep{chen2020simswap}&   93.01 & 1.53 & 2.84 &  7.48   & 27.1 \\
BlendFace$^{\dagger}$~\citep{shiohara2023blendface}&   97.02 & 3.07 &  2.14 & 3.84  & 24.7 \\
HifiFace$^{\dagger}$~\citep{wang2021hififace}&   98.01 & 2.84 &2.51 & 10.25   & 22.3  \\
FaceDancer$^{\dagger}$~\citep{rosberg2023facedancer}&   \cellcolor[HTML]{D7FCD7}\textbf{98.84} & 2.04  & 7.97 & 16.30  & 78.3 \\
DiffSwap$^{\dagger}$~\citep{zhao2023diffswap}&   98.54 & 2.45 & 5.35 & \cellcolor[HTML]{D7FCD7}\textbf{2.16} &  46245.2  \\
\midrule
AlphaFace (Our) & 98.77 &  \cellcolor[HTML]{D7FCD7}\textbf{1.24} & \cellcolor[HTML]{D7FCD7}\textbf{2.03} &  2.71   & 24.1 \\
\bottomrule
\end{tabular}
}
\caption{Quantitative examples of on the FF++ dataset \citep{rossler2019faceforensics++}. $^{\dagger}$ denotes that the results were obtained from their source codes. \colorbox[HTML]{D7FCD7}{\textbf{Green}} highlights the best performances.}
\label{Table:ff_results}
\vspace{-2ex}
\end{table}

\begin{table}[!t]
\centering\resizebox{\linewidth}{!}{
\begin{tabular}{l|c|c|c|c}
\toprule
\textbf{Method} & \textbf{CSIM $\uparrow$ } & \textbf{Pose err $\downarrow$} & \textbf{Expr error $\downarrow$} & \textbf{FID $\downarrow$}\\
\midrule
FSGAN$^{\dagger}$~\citep{nirkin2019fsgan}&   0.105 & 5.31 & 4.02 & 43.64 \\
SimSwap$^{\dagger}$~\citep{chen2020simswap}&   0.180 & 3.92 & 3.81  & 16.89\\
BlendFace$^{\dagger}$~\citep{shiohara2023blendface}&   0.392 & 3.71 & 3.18 & 11.27\\
HifiFace$^{\dagger}$~\citep{wang2021hififace}&   0.092 & 5.01 & 4.65 & 12.68 \\
FaceDancer$^{\dagger}$~\citep{rosberg2023facedancer}&   0.401 & 4.72 & 3.31 & 10.54 \\
DiffSwap$^{\dagger}$~\citep{zhao2023diffswap}&   0.278 & 4.58 & 4.12 & 12.63\\
\midrule
AlphaFace (Ours) & \cellcolor[HTML]{D7FCD7}\textbf{0.471} &  \cellcolor[HTML]{D7FCD7}\textbf{2.97} & \cellcolor[HTML]{D7FCD7}\textbf{3.03} & \cellcolor[HTML]{D7FCD7}\textbf{7.78} \\
\bottomrule
\end{tabular}
}
\caption{Quantitative results on the MPIE dataset. $^{\dagger}$ denotes that we run officially released source codes to obtain the results. \colorbox[HTML]{D7FCD7}{\textbf{Green}} highlights the best performances.}
\label{Table:mpie_results}
\vspace{-3ex}
\end{table}

\begin{figure}[t]
\centering
\includegraphics[width=0.9\columnwidth]{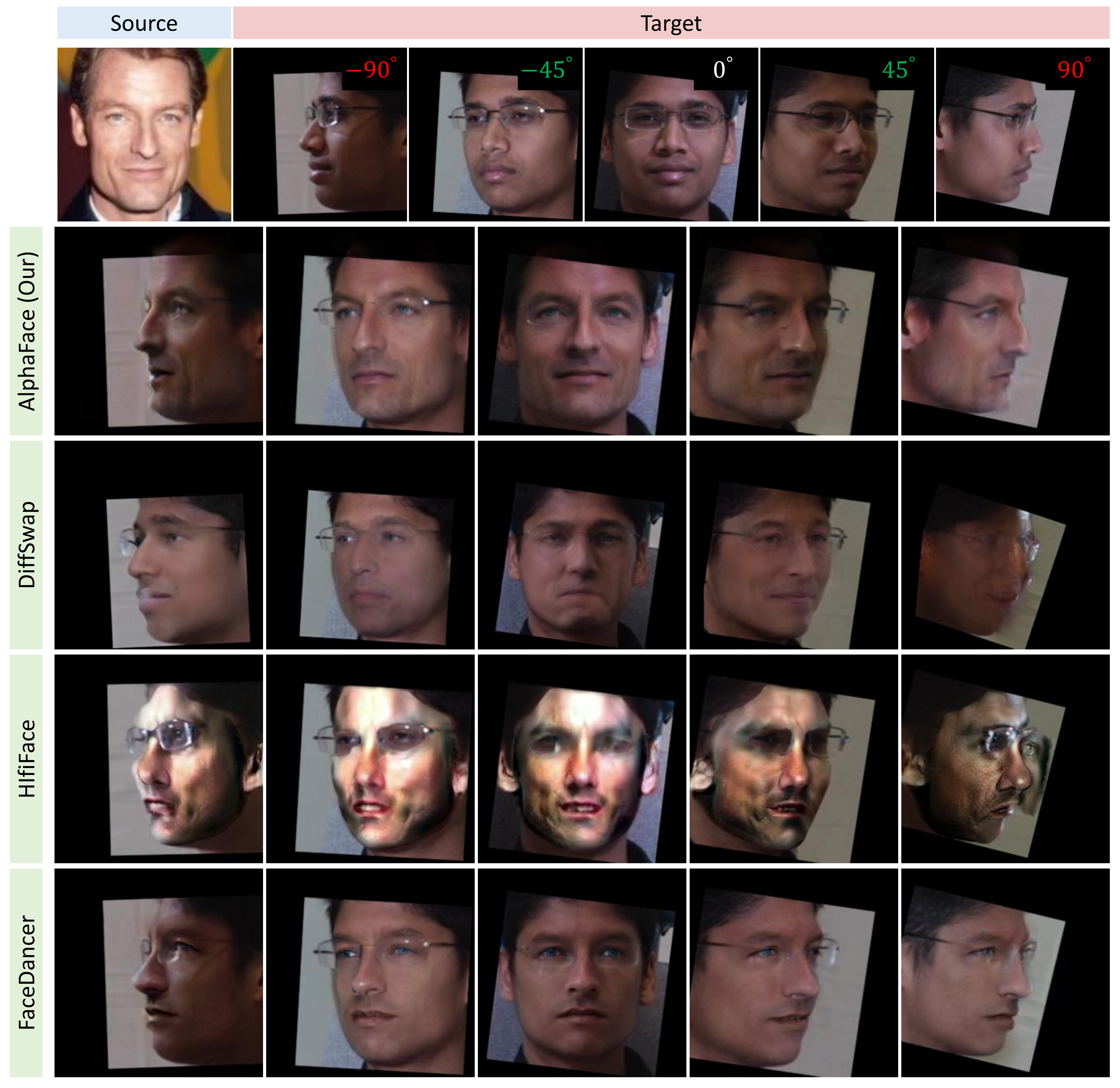}
\caption{\small The face identity swapping results of the AlphaFace and other methods  \citep{wang2021hififace,rosberg2023facedancer,zhao2023diffswap} on the MPIE dataset \citep{gross2010multi}. The extended results are shown in Appendix \ref{appx:mpie_results}.}
\label{fig:mpie_results}
\vspace{-3ex}
\end{figure}

\noindent
\textbf{Comparison on MPIE and LPFF:} Qualitative exemplars are provided in Fig. \ref{fig:mpie_results}, and quantitative comparisons appear in Table \ref{Table:mpie_results}. AlphaFace delivers the strongest overall performance, attaining the highest CSIM (0.471), the lowest pose (2.97) and expression (3.03) errors, and the lowest FID (7.78). By comparison, FaceDancer ranks second (0.401 CSIM, 4.72 pose and 3.31 expression errors, and 10.54 FID) with higher geometric discrepancies. At the same time, BlendFace yields 0.392 CSIM with 3.71 pose and 3.18 expression errors, trading slightly better alignment than FaceDancer for reduced identity similarity. These results indicate that AlphaFace preserves identity more faithfully whilst simultaneously maintaining target pose and expression with greater accuracy under extreme facial pose changes. Visual assessments on MPIE (Fig.~\ref{fig:mpie_results}) corroborate the metrics. Most SOTA methods mislocalise facial regions or fail to generate plausible swaps in extreme poses ($\pm45^{\circ}$), leading to conspicuous artefacts and texture distortions. HifiFace degrades further, with distorted facial components and boundaries. FaceDancer generates relatively cleaner imagery and, in some cases, approaches the visual quality of AlphaFace, yet both display silhouette blurring under severe head rotations. DiffSwap tends to exhibit low identity similarity and very blurry facial boundaries for extreme viewpoints ($\pm90^{\circ}$).

LPFF results follow similar trends, where AlphaFace yields remarkably coherent faces than FSGAN, BlendFace, and DiffSwap in rotated or tilted poses. Figure \ref{fig:lpff_results} shows face swapping results of AlphaFace and other SOTA methods on rotated (Fig. \ref{fig:lpff_results}(a)) and tilted (Fig. \ref{fig:lpff_results}(b)) facial poses. FaceDancer produces competitive results, but its fine textures, such as skin, are not as natural as those of AlphaFace. Also, DiffSwap's outcomes are not on par with FaceDancer's or even visibly competitive with SimSwap's in terms of identity and boundary fidelity.


Across those benchmarks, AlphaFace consistently leads in pose alignment and expression accuracy. For the identity retrieval score, FaceDancer outperforms our method on the FF++ dataset by 0.07, achieving 98.84, but the performance gap is very tight, and AlphaFace produces lower pose and expression errors and a lower FID. In addition, FaceDancer takes 78.3 ms per frame (12.8 FPS), which is not enough for real-time applications, while AlphaFace takes 24.1 ms (41.5 FPS). Those results establish AlphaFace as a reliable solution for face-swapping in settings with substantial pose variability and other adverse conditions.


\section{Conclusion}
\label{sec:5}
We have introduced AlphaFace, a real-time face-identity swapping framework trained with rich semantic information obtained by an open-source VLM. By aligning semantically rich text information obtained from a VLM, with visual features whilst reinforcing visual identity similarity using cross-adapted source identity code by the CAII, AlphaFace has achieved robust swaps under extreme poses and expressions without explicit geometric priors or auxiliary processing, sustaining about 40 FPS. Across three public benchmarks, AlphaFace has delivered competitive identity retention relative to the SOTA, achieved the best pose and expression errors, and maintained throughput comparable to real-time systems.


Blind spots remain. We tested several open-source VLMs and empirically selected OpenGVLab/InternVL3-14B based on empirical results. Our current work lacks an in-depth analysis of their use. Our future work focuses on in-depth analysis of VLM captions. It will include an ablation study on caption noise and the prompt, such as prompts for pose-only, expression-only, and pose-expression. 

{
    \small
    \bibliographystyle{ieeenat_fullname}
    \bibliography{main}
}

\clearpage
\appendix
\counterwithin{figure}{section}
\counterwithin{table}{section}

\section{Details of the datasets}
\label{appx:datasets}
We use five publicly available datasets to demonstrate the effectiveness of our AlphaFace VGGFace2~\citep{cao2018vggface2} dataset,  CelebA-HQ~\citep{karras2018progressive} dataset, FF++ dataset~\citep{rossler2019faceforensics++}, MPIE dataset~\citep{gross2010multi}, and LPFF dataset~\citep{wu2023lpff} are selected. Table \ref{appx:table:db_urls} shows the URLs to download the datasets that we used for this paper. The detailed information of the five datasets is as follows:

\begin{itemize}
\item \textbf{VGGFace2} contains 3.31 million images of 9131 subjects (identities), with an average of 362.6 images for each subject. Images are downloaded from Google Image Search and have large variations in pose, age, illumination, ethnicity and profession (e.g. actors, athletes, politicians). The whole dataset is split to a training set (including 8631 identites) and a test set (including 500 identites).

\item \textbf{CelebA-HQ} is a visually enhanced version of the CelebFaces Attributes dataset (CelebA)~\citep{liu2015faceattributes}, and it provides 30,000 images with 1024 $\times$ 1024 resolution. 

\item \textbf{FF++} is a forensics dataset consisting of 1000 original video sequences that have been manipulated with four automated face manipulation methods: Deepfakes, Face2Face, FaceSwap and NeuralTextures. The data has been sourced from 977 youtube videos and all videos contain a trackable mostly frontal face without occlusions which enables automated tampering methods to generate realistic forgeries. As we provide binary masks the data can be used for image and video classification as well as segmentation. In addition, we provide 1000 Deepfakes models to generate and augment new data.

\item \textbf{LPFF} comprises 19,590 high-quality, numerous identities, and extensive-pose diversity images. They firstly collect 155,720 raw portrait images from Flickr, then they remove all the raw images that have already appeared in FFHQ \citep{kazemi2014one}. After that, they align the remaining facial images and remove low-resolution images as well as noisy and blurred images.

\item \textbf{MPIE} contains over 750,000 images of 337 individuals. Each subject was photographed under 15 poses and 19 illumination conditions while exhibiting a range of facial expressions.
\end{itemize}

The data-preprocessing protocol is as follows: We firstly detect facial bounding boxes with YOLO5Face \citep{YOLO5Face} and align them by five-point landmark alignment following Bulat \etal\ \citep{bulat2017far}. After that, we regularise the image resolution to $256\times{}256$ to remove resolution variance. The source face images destined for the identity encoder are further down-sampled to $112\times{}112$ for matching with the input dimensionality of Arcface \citep{deng2019arcface}.

\begin{table}[h]
\centering\resizebox{\linewidth}{!}{
\begin{tabular}{l|l}
\toprule
\textbf{Dataset} & \textbf{Dataset URLs} \\
\midrule
VGGFace2~\citep{cao2018vggface2}&   \url{https://www.robots.ox.ac.uk/~vgg/data/vgg_face2}  \\
\midrule
CelebA-HQ~\citep{karras2018progressive} &   \url{https://mmlab.ie.cuhk.edu.hk/projects/CelebA}  \\
\midrule
FF++~\citep{rossler2019faceforensics++} &   \url{https://github.com/ondyari/FaceForensics}  \\
\midrule
LPFF~\citep{wu2023lpff} &   \url{https://github.com/oneThousand1000/LPFF-dataset} \\
\midrule
MPIE~\citep{gross2010multi}&   \url{https://www.kaggle.com/datasets/aliates/multi-pie} \\
\bottomrule
\end{tabular}
}
\caption{The URLs of the datasets used for this paper.}
\label{appx:table:db_urls}
\vspace{-2ex}
\end{table}

\section{List of public repositories.}
\label{appx:repositories}
We provide URLs of the public repositories of the methods that we selected for the experiment for the performance comparison on extreme face pose cases. Table \ref{appx:table:methods} shows the list of the public repositories.

\begin{table}[h]
\centering\resizebox{\linewidth}{!}{
\begin{tabular}{l|l}
\toprule
\textbf{Method} & \textbf{Public repository} \\
\midrule
FSGAN~\citep{nirkin2019fsgan} &   \url{https://github.com/YuvalNirkin/fsgan}   \\
\midrule
SimSwap~\citep{chen2020simswap} &   \url{https://github.com/neuralchen/SimSwap}  \\
\midrule
BlendFace~\citep{shiohara2023blendface}&   \url{https://github.com/mapooon/BlendFace} \\
\midrule
HifiFace~\citep{wang2021hififace}&   \url{https://github.com/maum-ai/hififace} \\
\midrule
FaceDancer~\citep{rosberg2023facedancer}&   \url{https://github.com/felixrosberg/FaceDancer} \\
\midrule
DiffSwap~\citep{zhao2023diffswap}&   \url{https://github.com/wl-zhao/DiffSwap} \\
\bottomrule
\end{tabular}
}
\caption{List of public repositories that can access the methods used for comparing face identity swap performances.}
\label{appx:table:methods}
\vspace{-2ex}
\end{table}

\clearpage
\onecolumn

\section{Example of a pair of a face image and the corresponding text description}
\label{appx:pair}
Table \ref{tbl:examplars} shows the examples of the training samples consisting of the facial image $x_{\text{t}}$, the corresponding facial segmentation mask $m_{\text{t}}$, and the corresponding text description $t_{\text{t}}$. As mentioned in the main manuscript, $t_{\text{t}}$ is obtained using OpenGVLab/InternVL3-14B \citep{chen2024internvl}, an open-source large-scale vision language model, with the prompt \textit{``Describe pose, background, facial accessories, and all obstacles covering the face area in the given face image. Only 70 words are allowed."}.

\begin{table*}[h]
\caption{Examples of a pair of a face image and a text description obtained by OpenGVLab/InternVL3-14B \citep{chen2024internvl}}
\label{tbl:examplars}
\centering
\setlength{\leftmargini}{0.4cm}
\resizebox{\textwidth}{!} 
{ 
\begin{tabular}{c|c|m{15cm}|}
\toprule
\toprule
Image $x_{\text{t}}$  & Mask $m_{\text{t}}$  & Description $t_{\text{t}}$ \\
\midrule
 \adjustbox{valign=c}{\includegraphics[width=20mm, height=20mm]{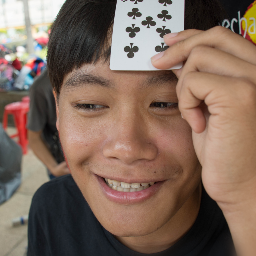}}   &  \adjustbox{valign=c}{\includegraphics[width=20mm, height=20mm]{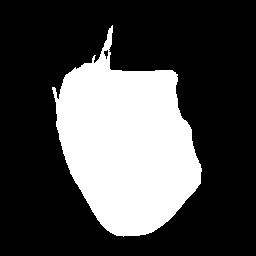}}   &   The person is smiling and holding a playing card with clovers in front of their forehead. The background is slightly blurred, showing an outdoor setting with people and red chairs. There are no facial accessories, but the card partially obscures the forehead and eyes.    \\ \midrule
\adjustbox{valign=c}{\includegraphics[width=20mm, height=20mm]{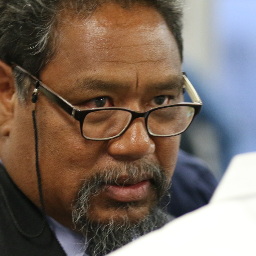}}   &  \adjustbox{valign=c}{\includegraphics[width=20mm, height=20mm]{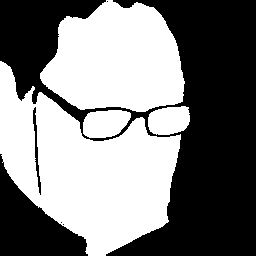}}   &   The person is looking downward, possibly at a document. The background is blurred with indistinct blue tones. They are wearing glasses and have a beard and moustache. There are no significant occlusions on the face.   \\ \midrule
\adjustbox{valign=c}{\includegraphics[width=20mm, height=20mm]{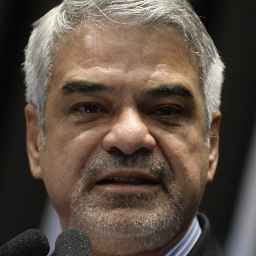}}   &  \adjustbox{valign=c}{\includegraphics[width=20mm, height=20mm]{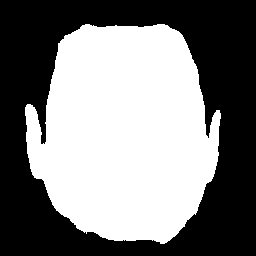}}   &   The person is facing forward with a neutral expression. The background is blurred and dark. There are microphones in front of the person, partially obscuring the lower part of the face. No facial accessories are visible. The individual has greying hair and a beard.   \\ \midrule
\adjustbox{valign=c}{\includegraphics[width=20mm, height=20mm]{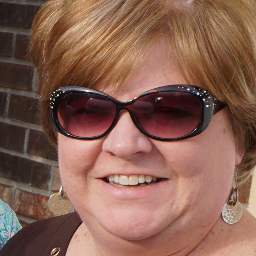}}   &  \adjustbox{valign=c}{\includegraphics[width=20mm, height=20mm]{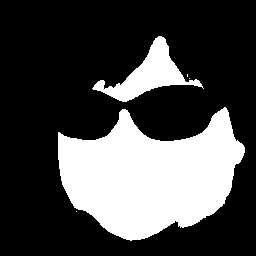}}   &  The person is facing the camera with a slight smile. The background features a brick wall. They are wearing large, round, studded sunglasses and dangling earrings. The sunglasses partially obscure the eyes, and the angle of the photo slightly covers the top of the head.   \\ \midrule
\adjustbox{valign=c}{\includegraphics[width=20mm, height=20mm]{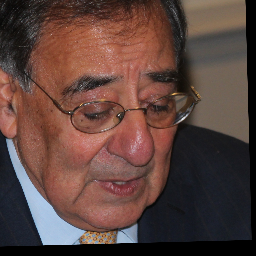}}   &  \adjustbox{valign=c}{\includegraphics[width=20mm, height=20mm]{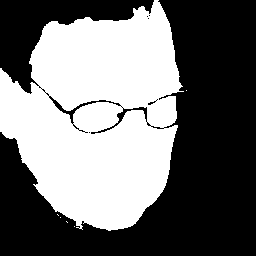}}   &  The person is wearing glasses and a suit with a patterned tie. They are looking downward, possibly reading or examining something. The background is blurred and indistinct, suggesting an indoor setting. There are no significant occlusions on the face, allowing clear visibility of facial features.  \\ \midrule
\adjustbox{valign=c}{\includegraphics[width=20mm, height=20mm]{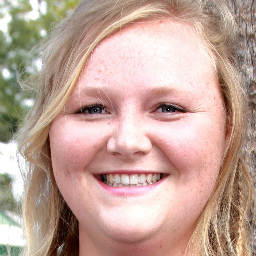}}   &  \adjustbox{valign=c}{\includegraphics[width=20mm, height=20mm]{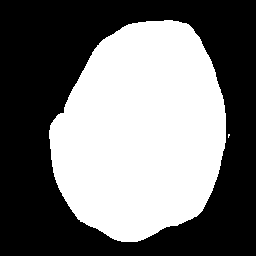}}   &  The person is facing the camera with a slight head tilt, smiling. The background includes trees and a blurred structure, suggesting an outdoor setting. There are no facial accessories. The right side of the face is partially obscured by a tree trunk. The lighting is natural, highlighting the person's features.  \\ \midrule
\adjustbox{valign=c}{\includegraphics[width=20mm, height=20mm]{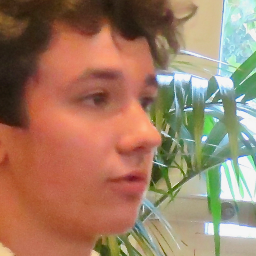}}   &  \adjustbox{valign=c}{\includegraphics[width=20mm, height=20mm]{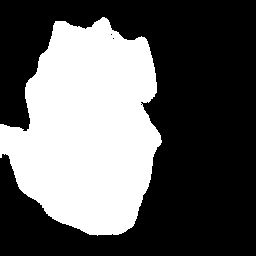}}   &  The person is facing slightly to the right with a neutral expression. The background includes a green plant with long leaves and a window. There are no facial accessories. The lighting is natural, and there are no significant occlusions on the face.  \\ \midrule
\adjustbox{valign=c}{\includegraphics[width=20mm, height=20mm]{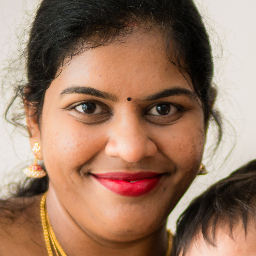}}   &  \adjustbox{valign=c}{\includegraphics[width=20mm, height=20mm]{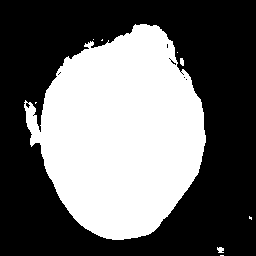}}   &  The person is facing the camera with a slight smile. The background is plain and light-colored. They are wearing gold earrings and a necklace. A bindi is on their forehead. Part of another person's head is visible in the lower right corner, partially obscuring the face. \\ \midrule
\adjustbox{valign=c}{\includegraphics[width=20mm, height=20mm]{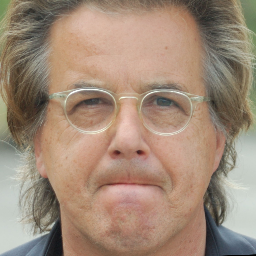}}   &  \adjustbox{valign=c}{\includegraphics[width=20mm, height=20mm]{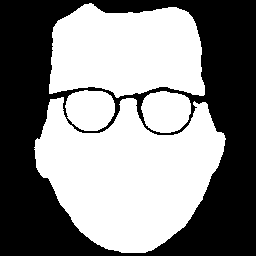}}   & The person is facing forward with a neutral expression. The background is blurred and indistinct. They are wearing round, thin-framed glasses. There are no significant occlusions on the face. The hair is medium-length and slightly tousled.  \\ 
\bottomrule
\bottomrule
\end{tabular}
}
\vspace{-2ex}
\end{table*}
\textbf{}

\clearpage
\onecolumn
\section{Qualitative results for Ablation study on identity injection}
\label{appx:abl_connection}
The CAII conducts cross-adaptation between the target latent features and the source identity code, applying adaptive instance normalisation (AdaIN) to both. In particular, the target-to-source adaptation (Eq. \ref{eq:original_source}) is the major engineering difference compared with identity injection approaches applied to other face swapping methods \citep{cui2023face,bitouk2008face,shiohara2023blendface,li2019faceshifter,zhao2023diffswap,wang2021hififace,rosberg2023facedancer,chen2020simswap,chen2023simswap++}. The target-to-source adaptation is firstly conducted by AdaIN to the source identity code using the target latent features; after that, we compute the residual operation using the original source identity code with the output of the AdaIN. By combining the source identity code and an adapted identity code, we can obtain a more aligned source identity code which can represent source identity without degrading in describing target attributes.

\begin{figure}[h]
\centering
\includegraphics[width=\linewidth]{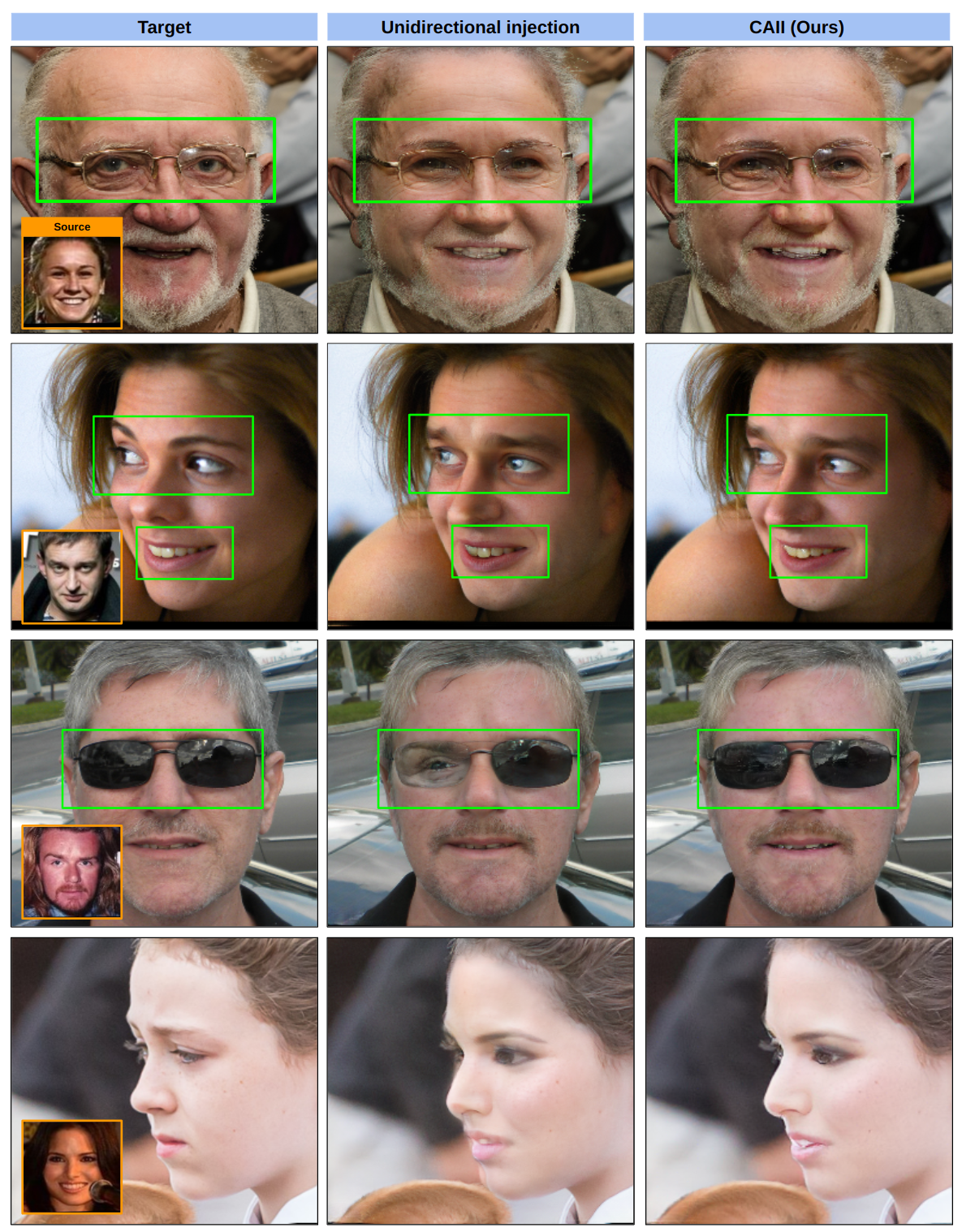}
\caption{\small  The face identity swapping results of the AlphaFace depending on the identity injection approaches.}
\label{fig:abl_fig2}
\vspace{-2ex}
\end{figure}

The qualitative results in Table \ref{table:abl2} demonstrate the effectiveness of the CAII. Additionally, we provide qualitative results. Figure \ref{fig:abl_fig2} shows the qualitative results for the ablation study to analyse the effectiveness of the cross-adaptive identity injection (CAII) block for face identity swapping. The areas that show obvious differences are highlighted in green boxes. As shown in Figure \ref{fig:abl_fig2}, the CAII encourage the generation of more detailed visual components in face identity swapping using AlphaFace. It is shown by more similar wrinkles, eye gaze, and successfully generated facial accessories. These results suggest the CAII-based AlphaFace achieves better quantitative results for pose and expression errors than the other. Also, in face-swapping under extreme facial poses, the results using the CAII show clearer facial boundaries than those using unidirectional injection, as evidenced by better FID scores.

\clearpage
\onecolumn
\section{Extended results on the FF++ dataset}
\label{appx:ff++_results}
\begin{figure}[h]
\centering
\includegraphics[width=0.9\linewidth]{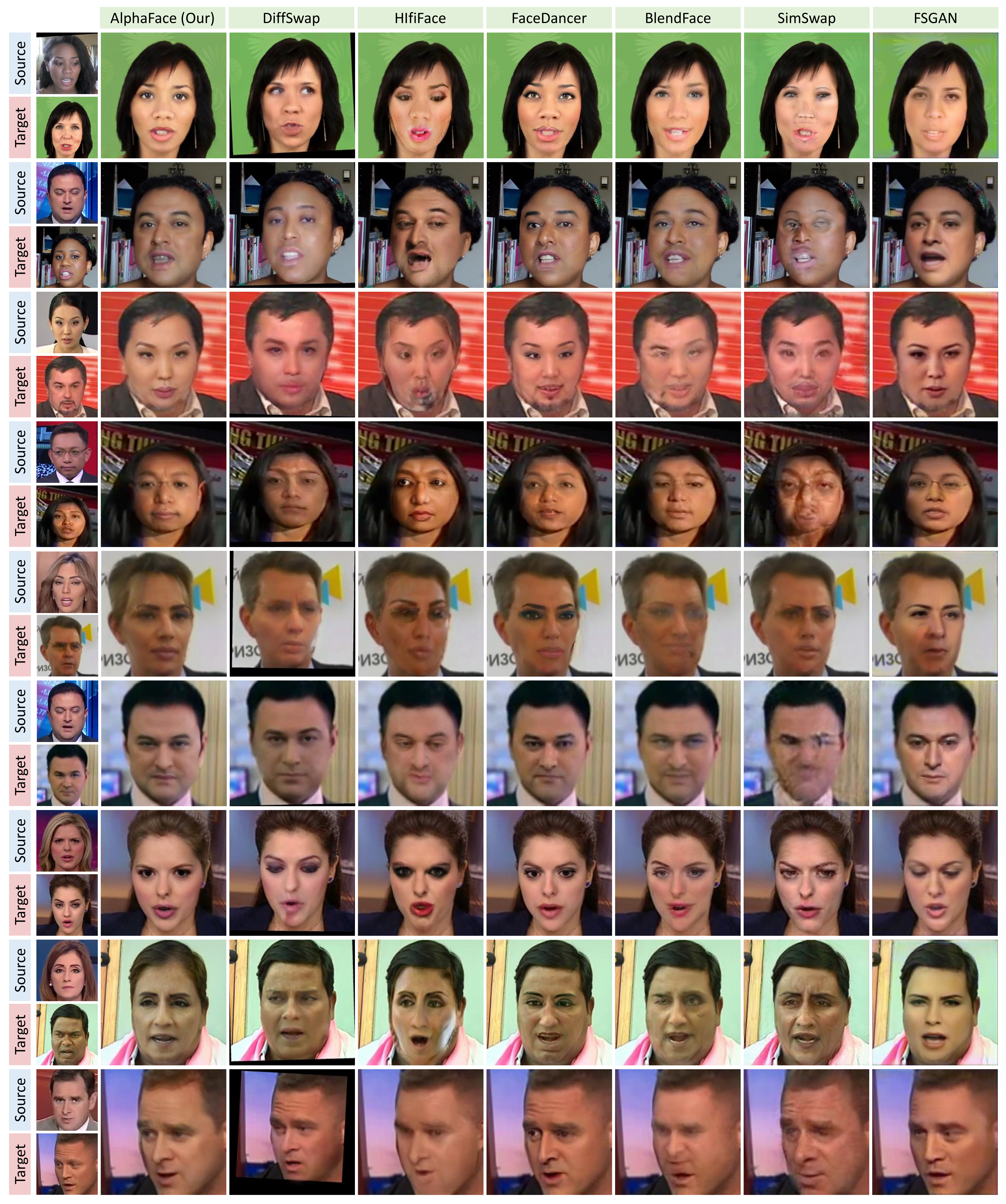}
\caption{\small  The face identity swapping result of the AlphaFace depends on the identity injection approaches.}
\label{fig:ff++_appx1}
\vspace{-2ex}
\end{figure}

Figure \ref{fig:ff++_appx1} shows the extended results of Figure \ref{fig:ff_qualtitative} for face identity swapping on the FF++ dataset. We can still observe that BlendFace generates some hallucinated faces that are totally mismatched with the actual face area. FSGAN results are significantly blurry, but also sometimes do not change much. HifiFace's results contain high contrast, making their swapped results totally disrupted. SimSwap achieves competitive performance; however, in extreme poses, the boundaries between the face and background are not clear enough. The face-swapping results using AlphaFace show the most natural results. 

\clearpage
\onecolumn

\section{Extended results on the MPIE dataset}
\label{appx:mpie_results}
\begin{figure}[h]
\centering
\begin{minipage}[b]{0.49\linewidth}
        \centering
        \includegraphics[width=\linewidth]{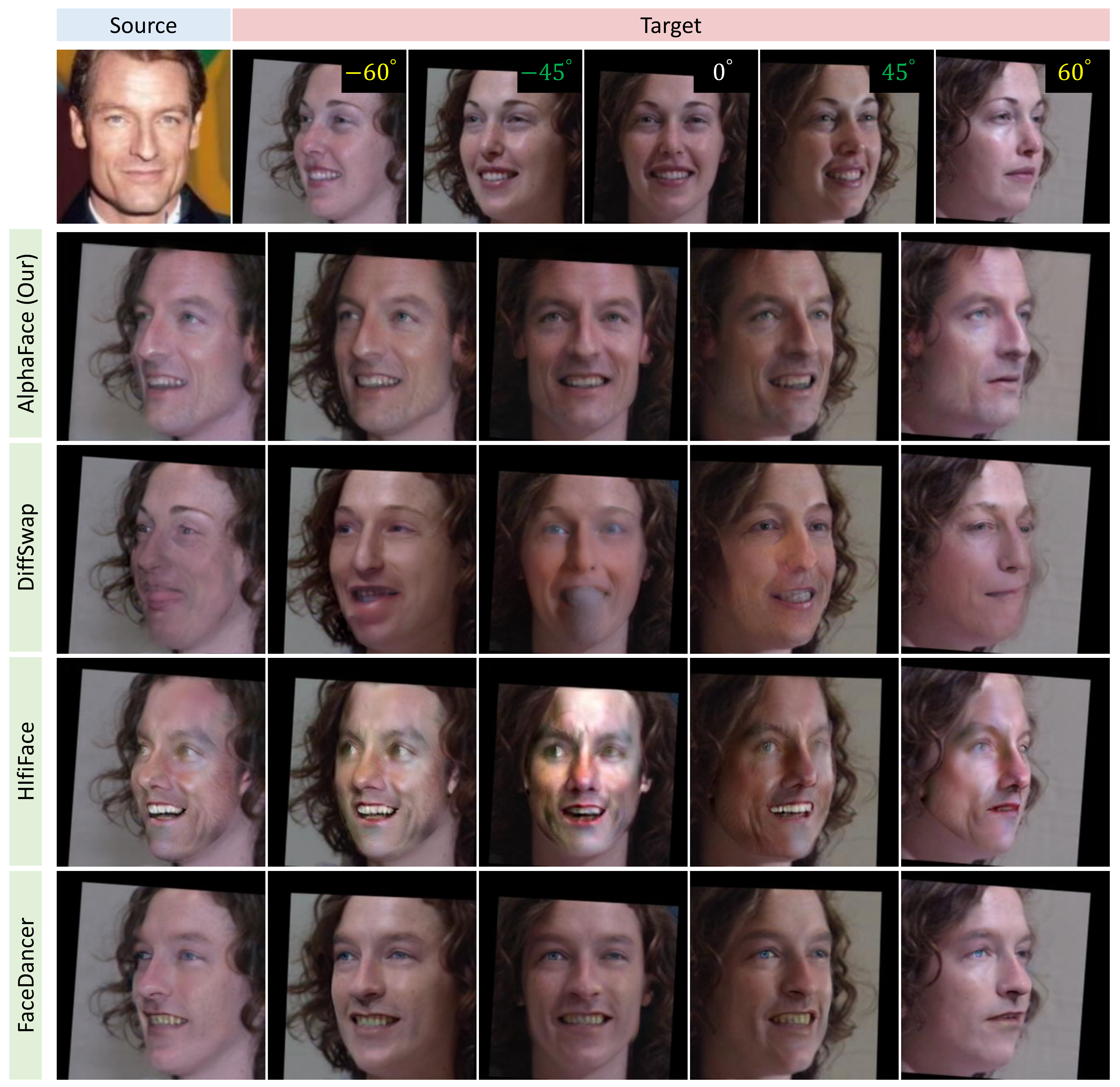}
        \phantomsubcaption \label{subfiglcr:left}
    \end{minipage}
    \hfill
    \begin{minipage}[b]{0.49\linewidth}
        \centering
        \includegraphics[width=\linewidth]{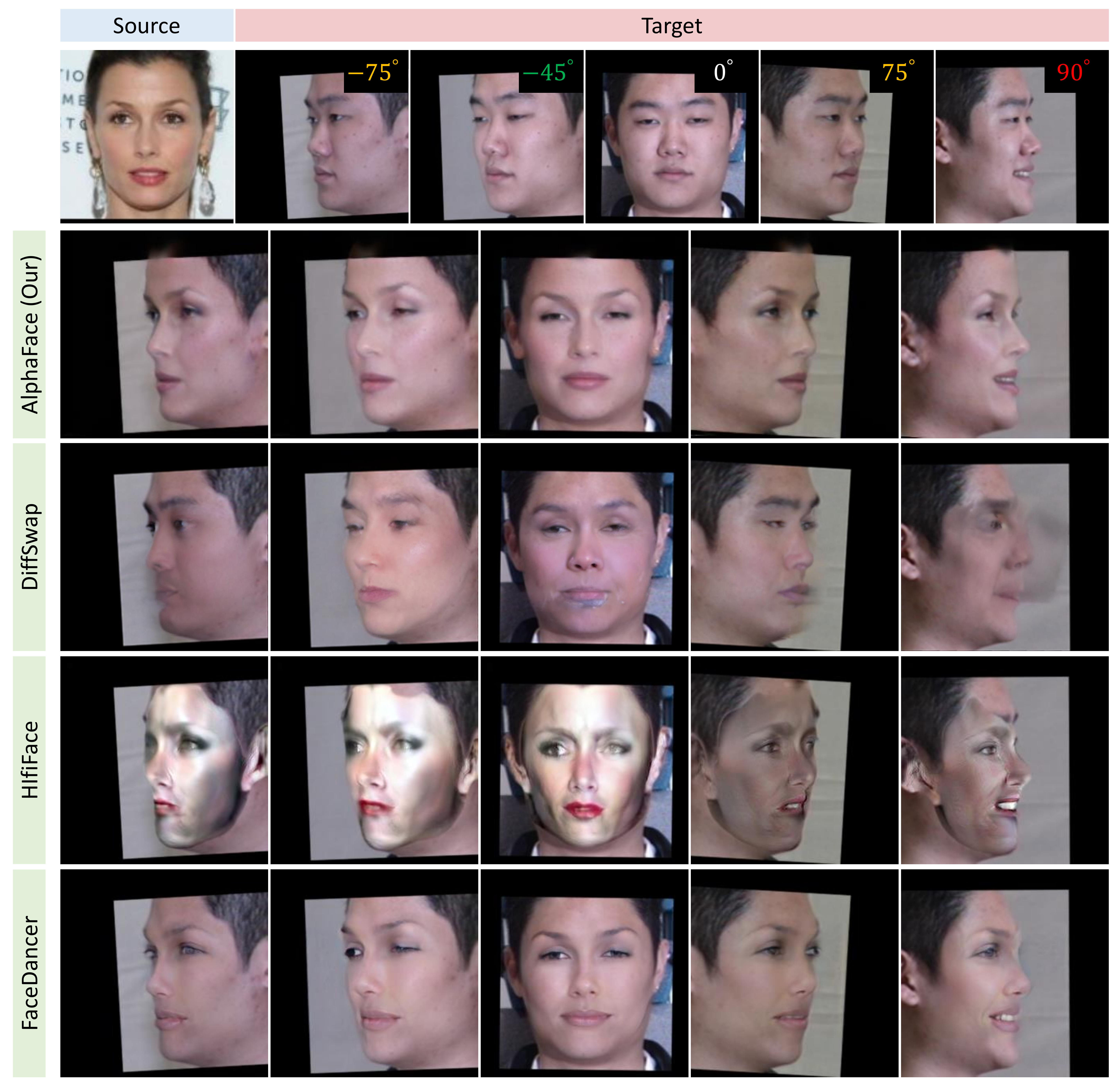}
        \phantomsubcaption \label{subfiglcr:right}
    \end{minipage}
\caption{\small  The face identity swapping result of the AlphaFace and other methods \citep{zhao2023diffswap,wang2021hififace,rosberg2023facedancer} on the MPIE dataset.}
\label{fig:mpie_appx1}
\vspace{-2ex}
\end{figure}

Figure \ref{fig:mpie_appx1} shows the extended results of Figure \ref{fig:mpie_results} for face identity swapping on the MPIE dataset. In this experiment, we only evaluate DiffSwap \citep{kim2025diffface}, HifiFace \citep{wang2021hififace}, and FaceDancer \citep{rosberg2023facedancer}. DiffSwap is a diffusion-based face swapping method, and the remaining two methods, in particular, aim to improve pose robustness by exploiting geometric features. The results of HifiFace and DiffSwap contain high contrast, making their swapped results totally disrupted. In particular, the results of HifiFace are totally distorted. FaceDancer outputs some competitive results with AlphaFace; however, in source identity representation, AlphaFace results show stronger source identity. Additionally, some facial components of the swapped faces based on FaceDancer are a little bit disrupted. 

\clearpage
\onecolumn

\section{Extended results on the LPFF dataset}
\label{appex:lpff_extended}
\begin{figure*}[h]
\centering
\includegraphics[width=\linewidth]{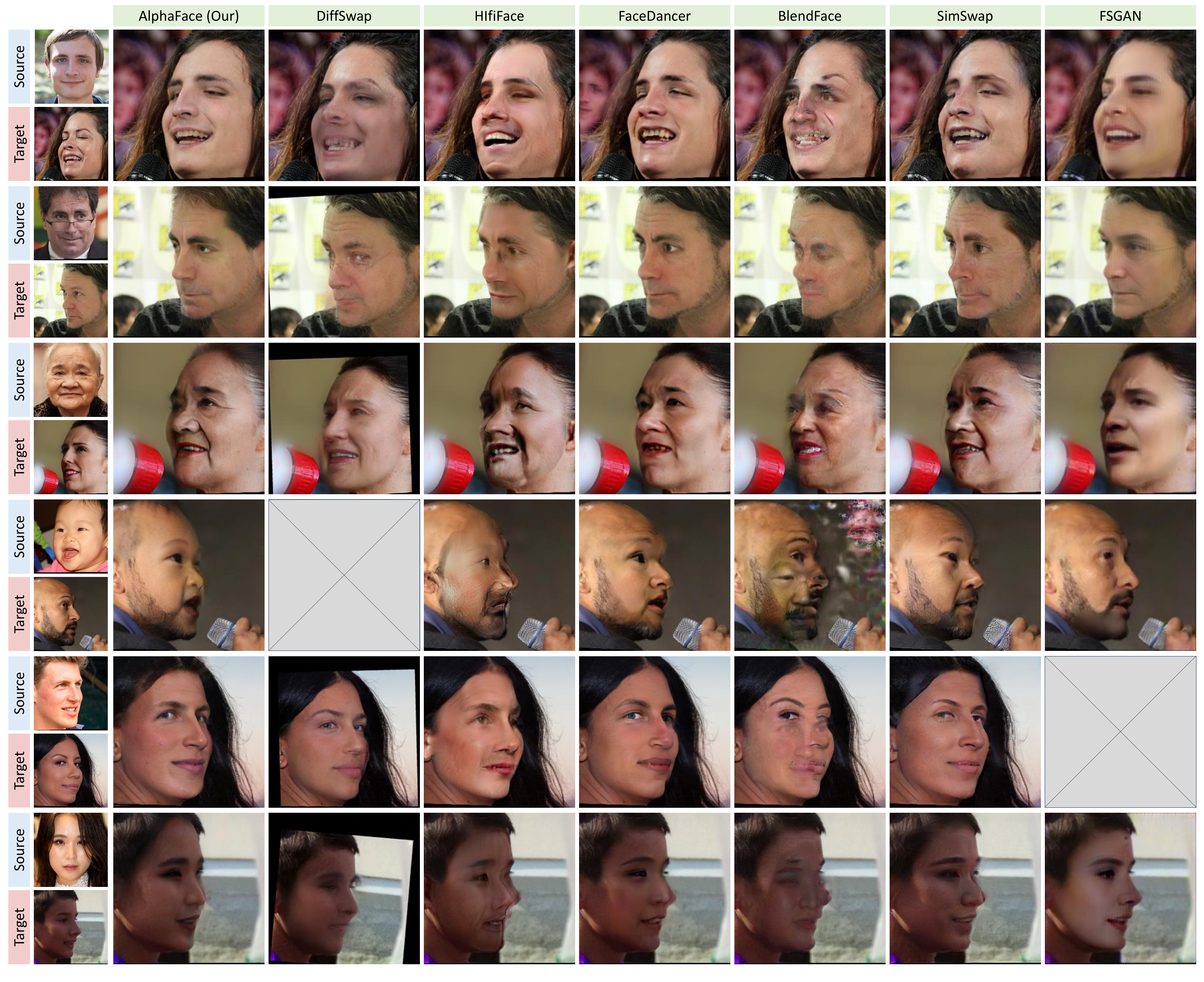}
\caption{\small  The face identity swapping result for horizontally rotated faces of the AlphaFace and other methods \citep{shiohara2023blendface,nirkin2019fsgan,wang2021hififace,chen2020simswap} on the LPFF dataset. The x-boxes indicate that a face-swapping method failed to generate the swapping results in some way, so there is no output.}
\label{fig:appx_lpff1}
\end{figure*}

\begin{figure*}[h]
\centering
\includegraphics[width=\linewidth]{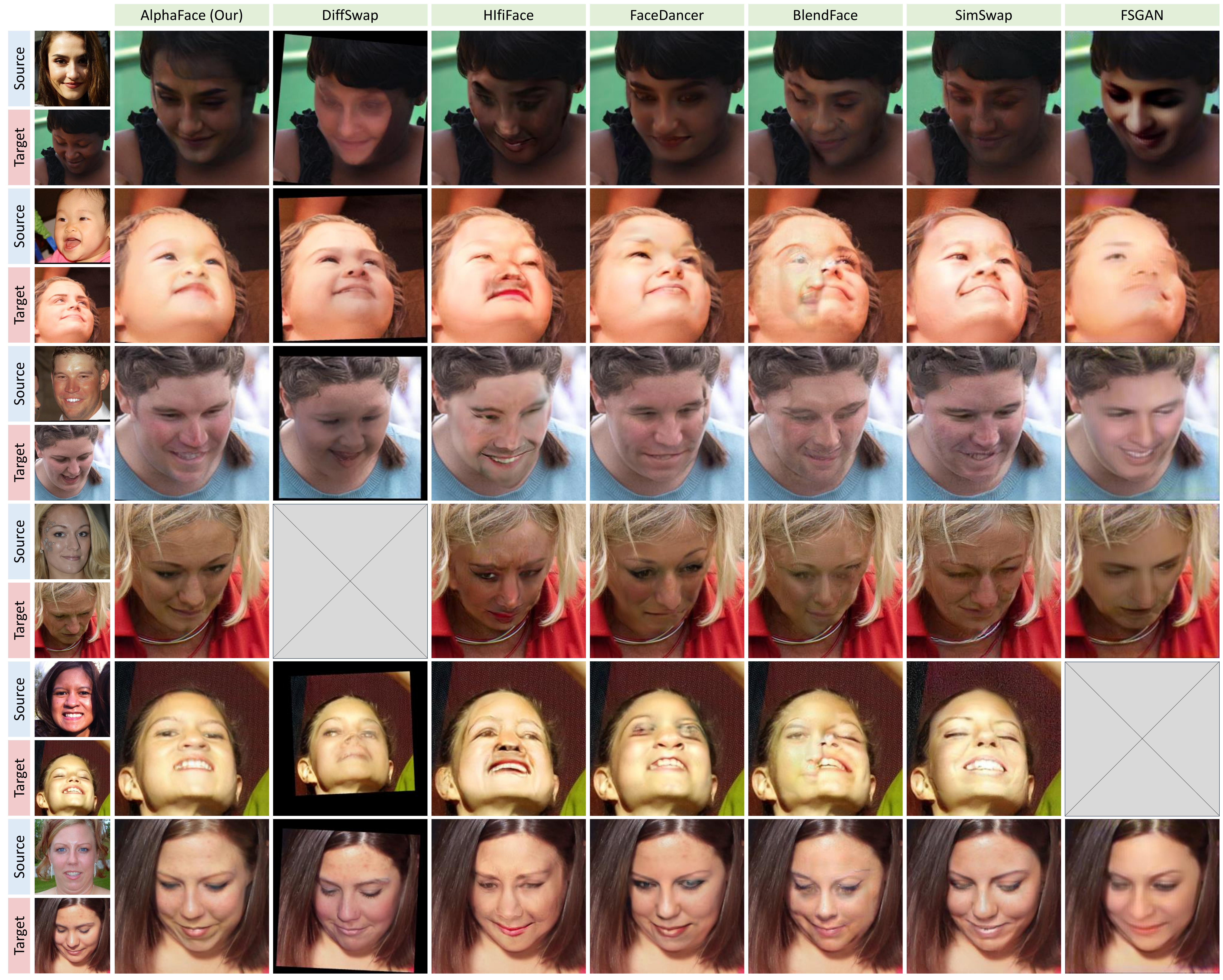}
\caption{\small  The face identity swapping result for vertically tilted faces of the AlphaFace and other methods \citep{shiohara2023blendface,nirkin2019fsgan,wang2021hififace,chen2020simswap} on the LPFF dataset. The x-boxes indicate that a face-swapping method failed to generate the swapping results in some way, so there is no output.}
\label{fig:appx_lpff2}
\end{figure*}

Figure \ref{fig:appx_lpff1} and Figure \ref{fig:appx_lpff2} show the extended results of Figure \ref{fig:lpff_results} for face identity swapping on the LPFF dataset. Figure \ref{fig:appx_lpff1} contains the face swapping results on rotated faces, and Figure \ref{fig:appx_lpff2} shows the results on tilted faces. 

\end{document}